\documentclass[10pt,journal,compsoc]{IEEEtran}
\ifCLASSOPTIONcompsoc
  \usepackage[nocompress]{cite}
\else
  \usepackage{cite}
\fi

\ifCLASSINFOpdf
   \usepackage[pdftex]{graphicx}
  \DeclareGraphicsExtensions{.pdf}
\else
   \usepackage[dvips]{graphicx}
  \DeclareGraphicsExtensions{.eps}
\fi

\usepackage{amsmath}
\usepackage{amssymb}
\usepackage{xcolor,colortbl}
\definecolor{LightCyan}{rgb}{0.88,1,1}
\usepackage{algorithmic}
\usepackage[ruled,vlined]{algorithm2e}

\usepackage{array}
\usepackage{mdwmath}
\usepackage{mdwtab}

\usepackage{eqparbox}

\begin{document}

\title{Binary Quadratic Programing for Online Tracking of Hundreds of People in Extremely Crowded Scenes}
\author{Afshin~Dehghan,~\IEEEmembership{Member,~IEEE,}
        and~Mubarak~Shah,~\IEEEmembership{Fellow,~IEEE}}

\markboth{Journal of \LaTeX\ Class Files,~Vol.~14, No.~8, August~2015}%
{Shell \MakeLowercase{\textit{et al.}}: Bare Advanced Demo of IEEEtran.cls for IEEE Computer Society Journals}

\IEEEtitleabstractindextext{%
\begin{abstract}
Multi-object tracking has been studied for decades. However, when it comes to tracking pedestrians in extremely crowded scenes, we are limited to only few works. This is an important problem which gives rise to several challenges. Pre-trained object detectors fail to localize targets in crowded sequences. This consequently limits the use of data-association based multi-target tracking methods which rely on the outcome of an object detector. Additionally, the small apparent target size makes it challenging to extract features to discriminate targets from their surroundings. Finally, the large number of targets greatly increases computational complexity which in turn makes it hard to extend existing multi-target tracking approaches to high-density crowd scenarios. In this paper, we propose a tracker that addresses the aforementioned problems and is capable of tracking hundreds of people efficiently. We formulate online crowd tracking as Binary Quadratic Programing. Our formulation employs target's individual information in the form of appearance and motion as well as contextual cues in the form of neighborhood motion, spatial proximity and grouping constraints, and solves detection and data association simultaneously. In order to solve the proposed quadratic optimization efficiently, where state-of art commercial quadratic programing solvers fail to find the answer in a reasonable amount of time, we propose to use the most recent version of the Modified Frank Wolfe algorithm, which takes advantage of SWAP-steps to speed up the optimization. We show that the proposed formulation can track hundreds of targets efficiently and improves state-of-art results by significant margins on eleven challenging high density crowd sequences.
\end{abstract}

}

\maketitle

\IEEEdisplaynontitleabstractindextext
\IEEEpeerreviewmaketitle

\ifCLASSOPTIONcompsoc
\IEEEraisesectionheading{\section{Introduction}\label{sec:introduction}}
\else
\section{Introduction}
\label{sec:introduction}
\fi

\IEEEPARstart{W}{hy} do we study crowds and why is tracking individuals in crowds important to us? The answer is safety. Of all the things which cause the eyes of the world to gaze upon the work of computer vision researchers, safety is the one which resonates most deeply as we as a society seek to prevent disasters and to protect individuals. When by dint of circumstance a large number of people move in a small area, safety becomes the biggest concern. Tragic incidents such as Boston marathon bombing, \cite{BostonMarathon} or the recent Hajj stampede \cite{HajjStampede} exemplify why there is a need for visual analysis of crowds. Moreover, understanding the dynamics of large groups of people is critical in the design and management of any type of public events. When dealing with high-density crowd scenarios such as religious rites participations, political rallies, concerts or marathons, modeling crowd dynamics can become quite complex. This could be due to several factors, including as the  heterogeneity of participants or their interactions with one and another.

While analyzing crowds, tracking individuals plays an important role and provides the prerequisite to many visual tasks such as crowd management, anomaly detection, activity recognition, crowd understanding and even crowd modeling for computer graphics. However, when looking at the literature, most multiple object tracking methods have focused on low- or medium-density crowd sequences \cite{KumarICCV2013,cmot,WenCVPR2014,ZamirGMCPECCV12}, where the number of targets are limited to tens of people. Almost none of these methods can be used to track people directly in crowds. This is either due to the complexity of their methods or the type of input data they require (e.g availability of detection candidates in each frame). 
\begin{figure}
	\centering
	\includegraphics[width=0.9\linewidth]{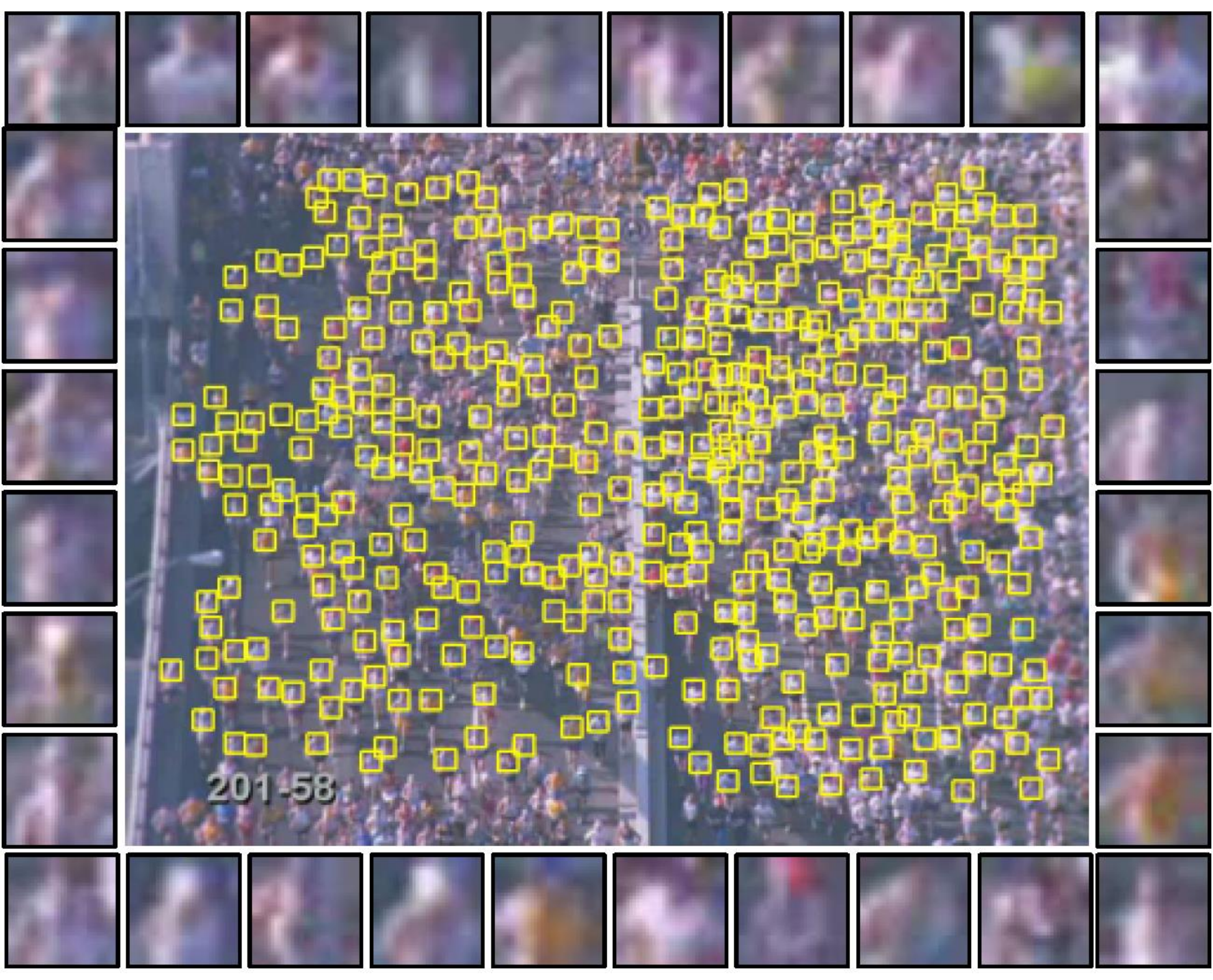}
	\caption{This figure shows one of our test sequences. The yellow boxes show the targets that are tracked in this sequence. On the borders we show the close-up versions of some of the targets. As can be seen targets are very small and the discriminative appearance cues are very low. Moreover targets look very similar which confuses most trackers. }
	\label{fig:Teaser}
\end{figure}

When dealing with high-density crowd scenes, containing hundreds of pedestrians, the problem becomes more difficult. It is even challenging, if not impossible, for humans to track individuals in such sequences (An example of such a sequence is shown in Figure \ref{fig:Teaser}). The first challenge that one would be faced with, while designing a tracker, is that pre-trained object detectors fail to detect individuals in high-density crowd scenes. The main reason is that most crowded sequences are captured using cameras facing down, where the full body is not visible. Moreover face/head detection methods have shown poor performances in such scenarios \cite{IdreesCVPR13}. This makes the use of data-association based tracking methods \cite{Nevatia2008,Hamed2011,Berclaz2011PAMI} impossible since they rely entirely on the outcome of a pre-trained object detector.

The other challenge in dealing with these sequences is, the small apparent target size. The number of pixels covering each target is small, which makes it hard to discriminate the target from others or sometimes from the background. Additionally, the large number of targets to be tracked increases the computational complexity and makes the design of an efficient tracker even more challenging. The latter issue, is one of the main reasons that all previous trackers \cite{IdreesIVC14, AliECCV2008, RodriguezCVPR2009, DineshCrowd2014}, focused on high-density crowds, track one target at a time instead of jointly optimizing the objective function for all the targets simultaneously. Although focusing on tracking one target at a time helps reduce the computational complexity, joint optimization is essential for optimal multi-target tracking. For example, multiple targets cannot occupy the same location at the same time. Thus for optimal assignments of new locations, objects in the scene should compete for each candidate location simultaneously, which is not the case in \cite{IdreesIVC14,AliECCV2008,DineshCrowd2014,RodriguezCVPR2009}. Additionally modeling interactions between targets can greatly benefit multi-target tracking algorithms. This is only feasible when target tracks are optimized jointly.

To this end, we propose a Binary Quadratic Programing solution to crowd-tracking that aims to accomplish the above-mentioned goals. To be more specific, we are the first to formulate tracking in high-density crowds as multi-target tracking. This means that our joint optimization allows updating tracks of all targets simultaneously. Further, our method considers multiple candidate locations for each target within the optimization and determines the correct location without assuming availability of target locations. Additionally, we propose five essential components for tracking individuals in crowds that capture target's individual information as well as contextual cues. We show that each of these components can be encoded in our objective function as a linear or a quadratic term. The first component captures the information necessary to discriminate each target from its background by using its appearance information. Our appearance term is based on an online discriminative learning approach, where we train a regression model for each target. This is different from previous works, where a generative model such as template based tracking is used as a baseline  \cite{IdreesIVC14, AliECCV2008, RodriguezCVPR2009, DineshCrowd2014}. The second and third components in our objective function capture the motion of the crowd. One encodes the target motion which is obtained based on the past trajectory of each target. The other one is related to the neighborhood motion, which captures the effect of neighbors. The fourth term in our objective function is the spatial proximity term that aims to discourage the co-selection of targets that are too close to each other. Finally, the last term encodes the group formation. It encourages the co-selection of targets that help maintain the group formations from frame to frame.

\begin{figure}
	\centering
	\includegraphics[width=0.9\linewidth]{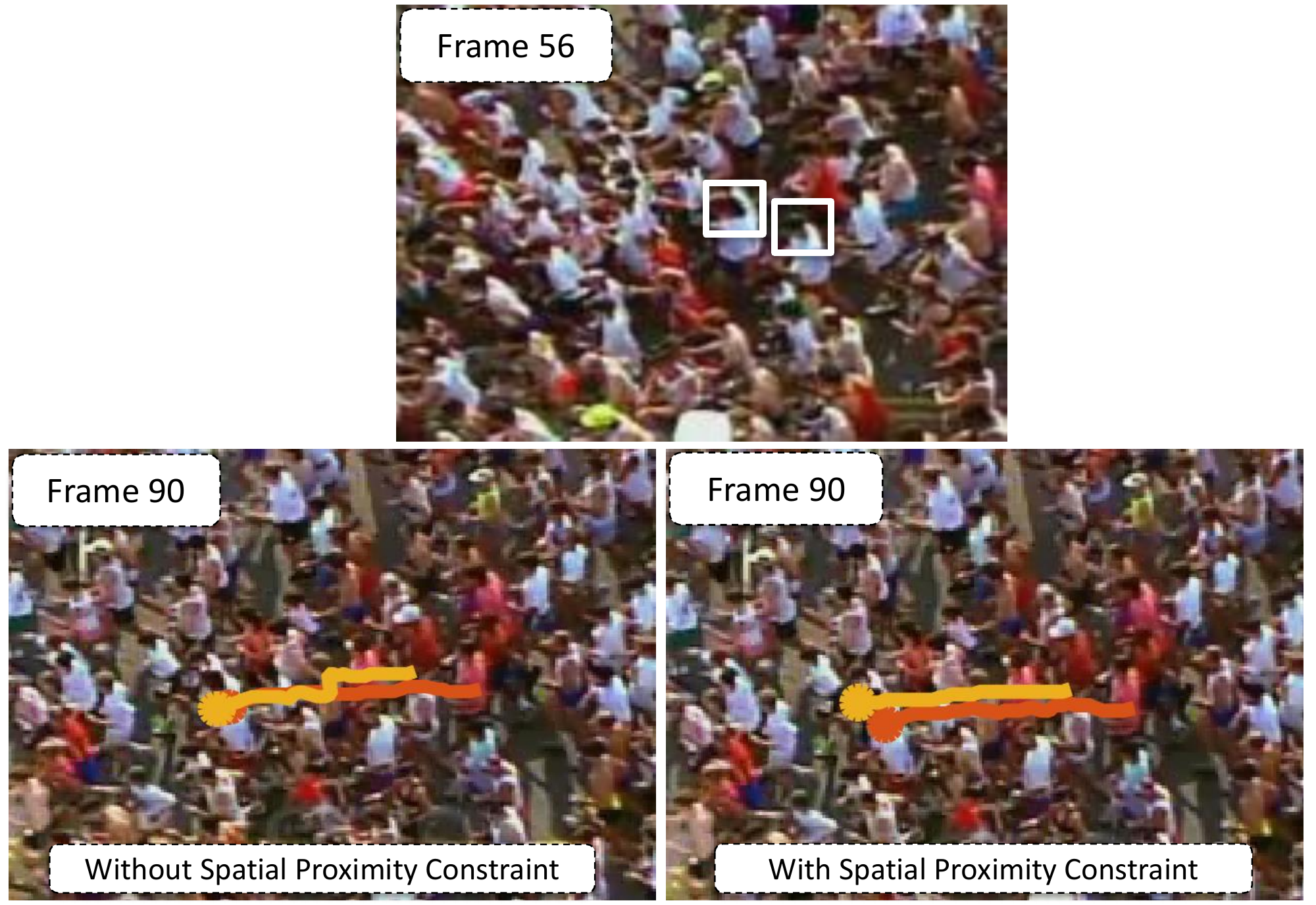}
	\caption{This figure motivates the spatial proximity constraint used in our formulation. The top figure shows the two targets with very similar appearance. The bottom figures demonstrate the tracking results of our method, with and without proximity constraint. As can be seen when no spatial proximity constraint is used, the tracker gets confused and tracks of one target jump to the near by one with similar appearance and motion. However, we are able to correctly track the two targets when we use the spatial proximity constraint as shown in bottom right figure.}
	\label{fig:motionvation_spatial}
\end{figure}

The first two terms are the building blocks of most current trackers. But the third and fourth terms are especially important for tracking targets in crowded scenes. When tracking targets in high-density crowds,  relying on only the observations from an individual's target tracks is not sufficient. In such scenarios, modeling contextual information and interactions between targets become vital. In crowded scenes, where a large number of people are bound to move in a small area, the motion of each individual is affected not only by its own behavior, but also by the motion of its neighbors. This neighborhood motion helps us to improve tracks of the targets when an individual's appearance and motion cues are not that strong.This happens frequently due to the low apparent target size and similarly looking moving targets. This neighborhood motion is captured through the third component of our objective function which encourages each individual to walk with similar motion as his/her coherently moving neighbors. The coherent motion groups are found using the information from past trajectories of targets. Our model is simpler compared to the previous works, which require prior information about the scene such as floor fields, \cite{AliECCV2008} or the heuristics such as instantaneous flow to deal with anomalies or unstructured scenes \cite{IdreesIVC14}. We discuss the drawbacks of these methods in detail in Section \ref{sec:RelatedWork}.

Another important component in our formulation which has not been used before in crowd-tracking is the spatial proximity constraint. The most commonly used constraint in almost all multi-target tracking methods is that each location should be assigned to only one target. While this constraint is essential, we found this not to be sufficient in crowded scenes. In data association based trackers, sparse detections are provided at input level, thus having the constraint that two tracks should not share a detection suffices. However, since we do not assume the availability of target detections, in every frame we have candidates sampled densely over the entire frame. Having targets with similar appearance and densely sampled candidates, it often happens that tracks of different targets overlap considerably while not sharing the same candidate location (i.e they may select two locations that are too close to each other). Additionally, when dealing with crowd sequences in aerial videos, having detections that overlap is restricted. This is because most sequences are captured using cameras facing down and targets do not occlude each other. Figure \ref{fig:motionvation_spatial} illustrates an example where two targets with similar appearance are running next to each other and the tracker gets confused after a few frames. In order to handle such failure cases, encouraging the co-selection of targets that are not too close to each other becomes essential. We later show in the experiment section that the spatial proximity constraint helps significantly improve the performance on most sequences. 

\begin{figure}
	\centering
	\includegraphics[width=1\linewidth]{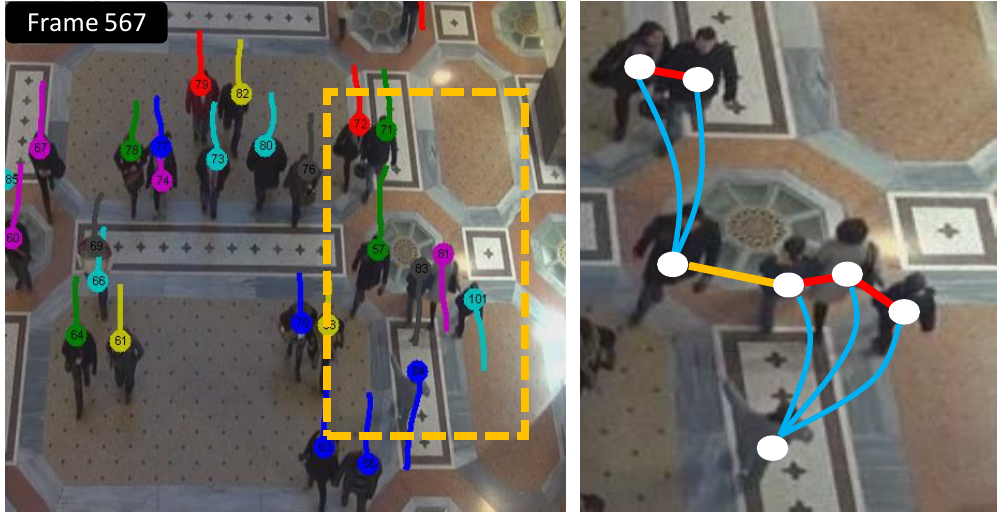}
	\caption{This figure is a graph illustration of the contextual constraints used in our formulation. The figure on the left shows the tracks and the figure on the right shows the constraints between the targets in yellow box. Each target is a node in the graph and all targets are connected with an edge. In practice there is a connection between every pair of targets, but here for simplicity we are showing only some of the connections. The figure illustrates two groups walking in opposite directions as well as two individuals.  Each edge is assigned different cost depending on the grouping information and distance of targets from each other.  The red edges that connect people in the same group encode the grouping and proximity constraints. The blue lines contain the neighborhood motion information and exist only between targets with coherent motion. The yellow edge only contains the spatial proximity information between two nearby targets. }
	\label{fig:method}
\end{figure}

In addition to our proximity constraint, we introduce another term which accounts for group formation. Group information has been recently integrated into several multi-target tracking algorithms and have shown its effectiveness \cite{group1,group2,group3}. Our Binary Quadratic Programing(BQP) formulation allows adding any constraint in the form of a linear or a quadratic term. We show that the group constraint can be formulated as an additional quadratic term and later show in our experiments that the group constraint helps to further improve the results. Please refer to Figure \ref{fig:method} for a summary of the constraints used in our formulation.

One of the main concerns for optimizing the proposed BQP is the number of variables, which correspond to potential locations that targets can occupy. Publicly available QP solvers such as CPLEX\cite{CPLEX} or MOSEK\cite{MOSEK}, can handle up to a few tens of targets. We show in our experiments that as the number of targets grows, the optimization becomes extremely inefficient. In some previous works, the quadratic function is converted into a linear one by adding additional equality/inequality constraints to the objective function \cite{SivicCVPR15}; one constraint for every pair of variables in the quadratic term. This may help reduce the complexity, but still it is not scalable to our problem size and it will not have the advantage of having a soft quadratic constraint in the objective function. Additionally, adding millions of constraints for large number of targets is not desirable. We instead use the modified Frank Wolfe algorithm with SWAP steps to directly solve the quadratic objective function. In our experiments, we show that the Frank-Wolfe with SWAP steps can handle sequences with up to a few hundreds of people. Lastly, we propose a straight-forward technique to reduce the number of candidates for further speed-up. We show that our speed-up technique reduces the computational complexity significantly, without loss in the tracking accuracy. 

In summary, the paper makes the following important contributions. We are the first ones to formulate tracking in high-density crowds as online multiple object tracking with joint optimization. We propose a flexible binary quadratic programing solution which brings in five essential components for crowd tracking into one single formulation. Our formulation includes both target's individual information as well as contextual cues and models the interaction between targets using neighborhood motion, spatial proximity and grouping constrains. We show that the proposed objective function can be solved efficiently using the most recent version of modified Frank-Wolfe algorithm. Additionally we propose an effective speed up technique that further reduces the computational complexity without loss in the tracking accuracy. Finally, we improve state-of-art on nine challenging sequences of \cite{IdreesIVC14} and two new sequences of Galleria1 and Galleria2.  

\section{Related Work}
\label{sec:RelatedWork}
Multiple target tracking is one of the fundamental problems in computer vision. Most prior works have focused on low and medium density crowd sequences \cite{KumarICCV2013,cmot,WenCVPR2014,MWISCVPR2011,CollinsCVPR2012, DehghanGMMCPCVPR15, ZamirGMCPECCV12}, where the goal is to design a better data association technique. Authors in \cite{MWISCVPR2011} formulate data association as maximum weight independent set. Many successful data association based trackers utilize network flow to formulate tracking \cite{Nevatia2008,Hamed2011,Berclaz2011PAMI}. The solution to network flow can be found efficiently using linear programing \cite{Berclaz2011PAMI} or a dynamic programing \cite{Hamed2011}. Authors in \cite{DehghanGMMCPCVPR15, ZamirGMCPECCV12, ristani2014tracking} formulate data association as maximum clique problem. All of these methods assume that, the detections in each frame are already given. This requires having a good pre-trained object detector \cite{DPM, DT05} that works reasonably well.  

When dealing with high-density crowds, the performance of pre-trained object detectors drop significantly. This is due to several factors. In crowded sequences, full bodies of people are not visible and only their heads, faces or upper bodies are visible. For instance, \cite{IdreesCVPR13} showed that the face or head detectors perform poorly in high density crowds. Instead of using a pre-trained object detector one can use an online object detector, which can be continuously updated. Due to this, there has been a recent interest in online tracking methods. In online tracking methods, pre-trained object detectors are not used and detection and data-association are solved at the same time. Online tracking methods have been used extensively in the context of single object tracking \cite{StruckICCV11, KCF, TLD}. However, there is only a handful of papers that use online tracking for multiple targets \cite{AfshinTINFCVPR15,SPOTCVPR13} and it is not easy to extend these methods to high-density crowd scenarios. 

Target tracking in highly crowded scene is relatively a new area of research, and only a handful of papers have focused on this problem \cite{IdreesIVC14, AliECCV2008, RodriguezCVPR2009, DineshCrowd2014}. The methods proposed in \cite{IdreesIVC14, AliECCV2008, RodriguezCVPR2009, DineshCrowd2014} track each target separately by training an online tracker for each individual separately. Ali and Shah \cite{AliECCV2008} proposed an algorithm which learns the prior information about the scene, called floor fields, that restrict the motion of each individual severely. This would cause failure in tracking when the crowd is dynamic, when there are anomalies or when camera moves. The series of papers by Kratz and Nishino \cite{KratzCVPR2010,KratzECCV2012,KratzPAMI2012} follow similar approach. They learn the motion patterns and then use them as prior information to improve track of each individual. Rodriguez et al. \cite{RodriguezCVPR2009} proposed a Correlated Topic Model to model crowd behavior at each location. Their model does not have the assumption of \cite{AliECCV2008} and allows targets to select among different motion patterns at each location. But still their model needs to learn dynamics of the scene and is prone to the same problems. In their approach, words correspond to low level quantized motion features and topics correspond to crowd behaviors. The recent work of \cite{IdreesIVC14} addresses some of the problems with previous works. In their approach, no prior assumption about the scene is used, instead the effect of neighborhood motion is incorporated in a greedy manner to improve tracking. Although the effect of neighbors in tracking was used before in\cite{PellegriniCVPR2009}, its extension to high-density crowds was explored first in \cite{IdreesIVC14}. 
 
One major draw-back of previous crowd trackers such as \cite{IdreesIVC14, AliECCV2008, RodriguezCVPR2009, DineshCrowd2014} is that, they track each target separately, lacking a joint optimization of target tracks. One of the main reasons for that is the complexity of joint optimization techniques. When dealing with hundreds of people, modeling interaction of targets becomes cumbersome and finding an efficient optimization is quite challenging. When tracking one target at a time, we are limited to use information from that target only. This makes it impossible to model interaction between the targets. In order to overcome this limitation, previous works have either used the prior information from the scene (which is not available and applicable all the time) or have tried to model interactions in a greedy manner by using information from other tracks \cite{IdreesIVC14}. Given the above issues, it is very crucial that multiple object tracking in crowded scenes should be formulated such that all target tracks are optimized simultaneously. This allows one to model the interactions between targets and include assumptions that are fundamental in modeling behavior of targets in physical world. In order to address the aforementioned problems, we present an online multi-object tracking framework where all the targets are tracked simultaneously. Our method provides a flexible formulation where one can include different individual and contextual terms.  

Our binary quadratic formulation consists of three linear terms and two quadratic term. Two linear terms capture the properties of the individual tracks. The third linear term as well as the quadratic terms are responsible for modeling interactions between the targets.  We show that the proposed quadratic objective function could be solved efficiently using an accelerated version of modified Frank-Wolfe algorithm which takes an advantage of \textit{SWAP} steps for further speed up\cite{svmFWIS14}. 

Frank-Wolfe optimization algorithm was introduced in $1956$ to solve the constraint quadratic programing, which has been recently revisited and used in many machine learning applications \cite{JaggiFWICML13,ssvmFWICML13,svmFWIS14}. Authors in \cite{ssvmFWICML13} used Frank-Wolfe with Away steps and proposed a faster optimization strategy for structure support vector machine. Allende et. al in \cite{svmFWIS14} showed that Frank-Wolfe could be applied to solve the quadratic programing in support vector machine and achieve significant speed up for large scale problems compared to its competitive methods such as projected gradient descend. Moreover, authors in \cite{JoulinVidLocECCV14} adopted Frank-Wolfe with away steps to solve the quadratic objective for image/video co-localization. In our work, we show that, commercial softwares such as CPLEX\cite{CPLEX} and MOSEK\cite{MOSEK}, which use Barrier Optimization techniques are not able to handle a large number of people. The accelerated FW that we use in our work not only can solve the problem efficiently for large number of target, but also its faster than commercial software and other version of Frank Wolfe used in \cite{JaggiFWICML13,ssvmFWICML13,JoulinVidLocECCV14}.

The rest of the paper is organized as follow, in Section. \ref{sec:frameWork} we present our framework. In Section \ref{sec:formulation}, we describe our binary quadratic formulation and different terms we consider in our objective function. The Frank Wolfe algorithm used for the optimization is covered in Section. \ref{sec:FW}. We describe our speed up technique in Section. \ref{sec:speedUP}. In Section. \ref{sec:experiments} we present our quantitative and qualitative results and finally in Section \ref{sec:conclusion} we conclude the paper.
 
\section{Proposed Framework}
\label{sec:frameWork}
We aim to solve the detection and data association in an online manner for high density crowd sequences. Given the initial target locations in the first frame, our method starts by training a discriminative model for each target using linear regression. During inference, the potential candidates for each target are sampled densely around the pervious locations of the target. Each candidate is assigned a cost according to its past trajectory as well as its surrounding neighbors. The goal is then to find new location of each target by minimizing the proposed quadratic objective function. The new target location is later used to update the target models if necessary.

\section{Objective Function}
\label{sec:formulation}

At every frame the best locations of the targets are found by minimizing the following objective function:
\begin{equation}
\begin{split}
\underset{\mathbf{x}}{\text{minimize}} f(\mathbf{x})=& \overbrace{\mathbf{c}_a^T\mathbf{x}}^{appearance}+ \overbrace{\zeta \ \mathbf{c}_m^T\mathbf{x}}^{target motion}+ \overbrace{\eta \ \mathbf{c}_{nm}^T\mathbf{x}}^{neighbor motion} \\ & +\overbrace{\mathbf{x}^T\mathbf{C}_{sp}\mathbf{x}}^{spatial proximity} +\overbrace{\mathbf{x}^T\mathbf{C}_{g}\mathbf{x}}^{grouping} , 
\end{split}
\label{eq:cost}
\end{equation}

\begin{figure}
	\centering
	\includegraphics[width=1\linewidth]{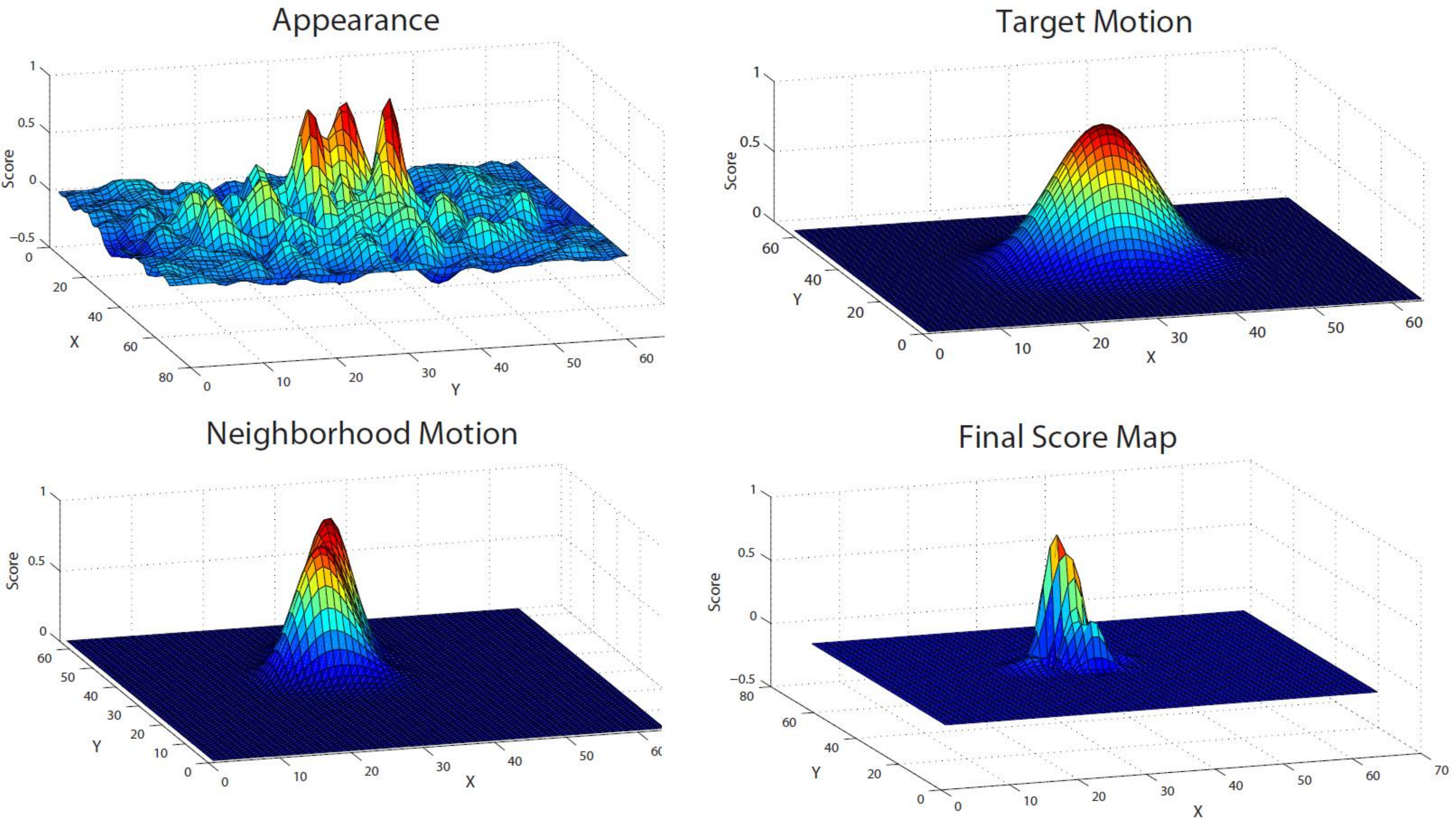}
	\caption{An example of the probability maps obtained for a target in seq-4. These maps include appearance, motion and neighborhood motion. It is clear from the figure that the appearance by itself is not always sufficient and the combination of all the three components is more discriminative for tracking.}
	\label{fig:linearTerms}
\end{figure}

\noindent where $\mathbf{x} \in \mathbb{R}^{l}$ is a vector which contains all the binary variables, each encoding potential location of the target in next frame. $l=n\times k$ defines the number of candidate locations, where $n$ is the number of targets and $k$ is the number of candidate locations for each target. $\mathbf{c}_a$, $\mathbf{c}_m$, $\mathbf{c}_{nm}$ are affinity vectors which respectively encode the appearance, motion and neighborhood motion cost. The affinity matrix $\mathbf{C}_{sp}$ includes the pairwise proximity cost that discourages the co-selection of locations that are too close to each others. This term is especially important in high density crowded scenes that are mostly captured by cameras facing down. Due to the camera view in these sequences targets will not occlude each other, thus two candidates cannot get closer than a certain distance to each other. The affinity matrix $\mathbf{C}_{g}$ contains the group information and encourages the people in the same group to keep their formation.\\
In order to ensure the solution found by solving Equation. \ref{eq:cost} is a feasible tracking solution, we need to enforce the following constrains:

\begin{equation}
\sum_{i \in N_j} x_i^j = 1, \hspace{0.1in} \{\forall i,j|1\leq i \leq k, 1\leq j \leq n \}
\label{eq:const1}
\end{equation}
\begin{equation}
x_i^j \in \{0,1\}, \hspace{0.1in} \{\forall i,j|1\leq i \leq k, 1\leq j \leq n \},
\label{eq:const2}
\end{equation}

\noindent where $x_i^j$, a component of vector $\mathbf{x}$, is a binary variable representing the $i^{th}$ candidate location in the neighborhood of the $j^{th}$ target. $k$ is the total number of sampled candidates for each target and $n$ is the total number of targets to track. The constraint in Eq. \ref{eq:const1} guarantees that exactly one location is selected for each target. The constraint in Eq. \ref{eq:const2} ensures that each location is assigned to at most one target. These constraints along with the cost function in Equation \ref{eq:cost} form our Binary Quadratic Programing formulation that needs to be solved at every frame. Each term in Equation. \ref{eq:cost} requires the computation of its own affinity matrix/vector. Below we describe in details how to compute these affinity matrices/vectors.

\textbf{Appearance} information in high density crowd sequence is not as discriminative as in low or medium dense crowd sequences such as the ones used in \cite{PET, ParkingLot}. An example is shown in Figure \ref{fig:Teaser}. There are only a small number of pixels covering each target, and the targets look very similar. However, our experiments show that the appearance still plays an important role in our approach. In \cite{IdreesIVC14,AliECCV2008,RodriguezCVPR2009}, a template-based method based on Normalized Cross Correlation (NCC) was used to capture such information. In a recent study in \cite{SmeuldersPAMI13}, it is shown that discriminative based tracking methods work better than the generative ones. However, the discriminative models have not been used in previous crowd tracking methods. One reason is the complexity of discriminative models. Training individual models, such as the one used in \cite{StruckICCV11}, for a large number of targets is computationally expensive. In this work, we show that one could still use discriminative models while not increasing the complexity. 

In our tracker, we train a regressor for each target by minimizing the following objective function:
\begin{equation}
\underset{\mathbf{w}_i}{\text{minimize}} \sum_{j \in T_i}{(\mathbf{w}_i \phi(x_i^j)-y_i^j)}^2+\lambda\left \| \mathbf{w}_i \right \|^2,
\label{eq:RidgeRegression}
\end{equation}

\noindent where $\mathbf{w_i}$ is the model parameters for the $i^{th}$ target, $\phi(x_i^j)$ is the feature vector extracted from the candidate location $x_i^j$, the labels are defined by $y_i^j$ and $T_i$ represents the training samples for $i^{th}$ target.  Regression models allow one to avoid binary labeling of training samples. This is shown in \cite{Struck} to provide a better model. The above optimization has a closed form solution of $\mathbf{w}=(Z^TZ+\lambda I)^{-1}Z^T\mathbf{y}$, where $Z$ is a matrix that has a sample $x_i^j$ per row. Once the models are trained the appearance cost for each candidate location is found using the following equation: 

\begin{equation}
c_{i,a}^j=\mathbf{w}_i\phi(x_i^j).
\label{eq:appearanceCost}
\end{equation}

In \cite{KCF} it is shown that the solution to Eq. \ref{eq:RidgeRegression} and \ref{eq:appearanceCost} can be found efficiently in Fourier domain. We follow the same approach to compute the models for each target, but instead of using gray scale images as used in \cite{KCF}, we employ the multi channel formulation of the above equation and use color features.

\textbf{Motion} plays an important role in the context of tracking pedestrians. The motion of pedestrian in crowded scene is effected by their environment. Thus using motion models that predict target location only based on its own behavior is not enough. Several methods have been proposed over the past few years to model the behavior of pedestrians in crowds considering their environment. But none of those models have been used in a crowd tracking framework with efficient joint optimization of target tracks. 

In this work, we use two different types of motion information. One that predicts target location based on its past observations, this is captured using $\mathbf{c}_m$. The other incorporates information from the neighboring targets \cite{IdreesIVC14} to predict the location of individuals at each time step.  

\begin{figure}
	\centering
	\includegraphics[width=1\linewidth]{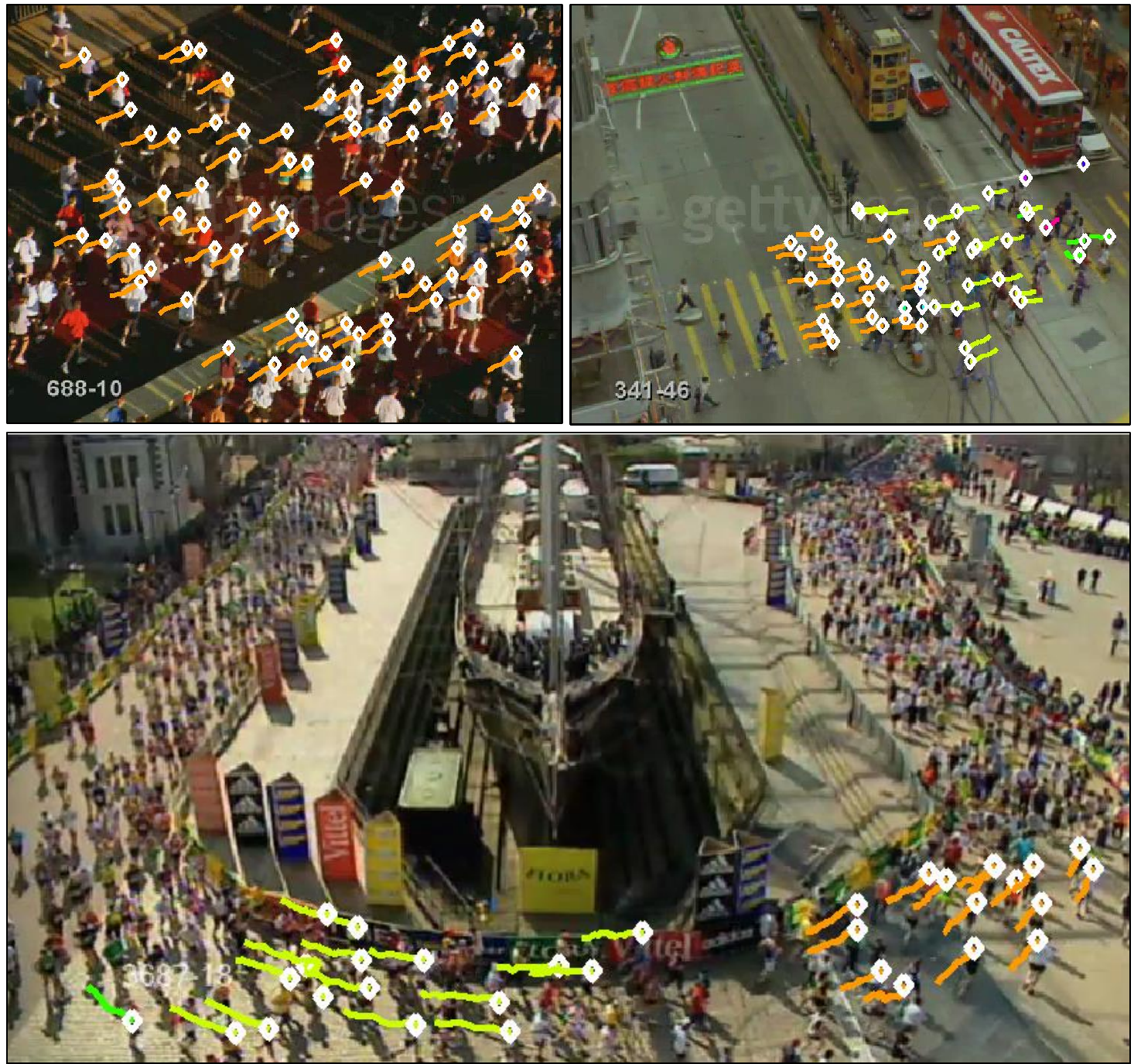}
	\caption{An example of groups that move with coherent motion in several sequences (each color corresponds to one group). These groups are used to incorporate the neighborhood motion effect in our optimization.}
	\label{fig:motionCoherent}
\end{figure}

The first term predicts the location of target based on its past observation using a linear velocity model shown below: 
\begin{equation}
\mathbf{c}_{i,m}=\mathcal{N}(A\mathbf{s}_i^{t-1} ,  \mathbf{\Sigma}),
\label{eq:targetMotionCost}
\end{equation}

\noindent where $\mathbf{s}_i^{t}=[p_i, \dot{p}_i]$ is a vector that contains the location and velocity of target $i$ at time $t$, $A$ is the state transition matrix and $\mathcal{N}(\mu, \Sigma)$ is a $2D$ Gaussian distribution. 

The second term, on the other hand, captures the local force that influence the motion of each individual. In order to capture the neighborhood motion, one needs to first find the groups of people with coherent motion. This is obtained by computing the correlation between the tracks based on their past observations (frames $t-10$ to $t-1$) and clustering them into different groups. An example is shown in Figure \ref{fig:motionCoherent}. During this unsupervised clustering we also take into account the distance between the members of the group. This approach is simple and effective and can be computed online without requiring much computations. Considering $p_i$ to be the neighbor of target $i$ which moves coherently with it. $N_i$ is a set that includes all the neighbors of target $i$ (In our experiments the number of neighbors in $N_i$ is limited to $5$). The influence of neighborhood motion is captured using the following equations.

\begin{equation}
c_{i,nm}=\sum_{j \in N_i}w_j.\mathcal{N} (A\mathbf{s}_{ij}^{t-1}, \mathbf{\Sigma}),
\label{eq:neighborMotionCost}
\end{equation}

\noindent where $\mathbf{s}_{ij}^{t}=[p_i, \dot{p}_j]$ encodes the position of target $i$ and velocity of the $j^{th}$ neighbor. The influence of each neighbor is captured using the weight coefficient $w_j$ which is obtained using the distance of the neighbor to the target:
  
\begin{equation}
w_j=\frac{exp(\left \| p_j-p_i \right \|_2)}{\sum_{k \in P_i}exp(-\left \| p_k-p_i \right \|_2)}.
\label{eq:neighborMotionCost}
\end{equation}

An example of the three linear terms for a target is shown in Figure \ref{fig:linearTerms}. It is clear from the figure that the appearance cue by itself is not always sufficient. One can clearly see that the combination of appearance, motion and neighborhood motion is able to discriminate target from its background better than using each feature individually. 

\textbf{Spatial Proximity Constraint}
is another important factor in our formulation. In top view high density crowd sequences, targets tend to look similar. An example is shown in Fig. \ref{fig:Teaser}. This similarity confuses most trackers, as the small apparent target size makes it difficult to extract useful appearance features. Our spatial proximity constraint discourages the tracker to select locations that are very close to each other. Additionally this is a soft constraint, meaning if the targets get too close (When the camera is not facing downward and we have a perspective effect) it still allows the targets to get close to each other as long as the appearance cues exist. 

In our formulation, $\mathbf{C}_{sp}$ contains the proximity cost for each pair, where $\mathbf{C}_{sp}=I-D^{-1/2}SD^{1/2}$ is the normalized Laplacian matrix \cite{graphcut}. $D$ is the diagonal matrix composed of row sums of similarity matrix $S$. $S \in \mathbb{R}^{l \times l}$ is the similarity matrix that encodes the spatial proximity cost and discourages the co-selection of locations that are too close (this is defined based on the target size). The entries of matrix $S$ are defined as follows: 

\begin{equation}
S_{ij}=exp(\frac{-\left \| p_i-p_j \right \|_2^2}{2\sigma^2}),
\label{eq:spatial constraint}
\end{equation}

\noindent where $\sigma$ is set to half the target size. It is important to note that we could not set $\mathbf{C}_{sp}=\mathbf{S}$. The reason is that matrix $\mathbf{S}$ is not positive-semi-definite and this makes our objective function non-convex. The Laplacian trick helps us to keep $\mathbf{C}_{sp}$ positive-semi-definite while still imposing the same effect to our optimization.

\textbf{Group Constraint} is another important factor that affects pedestrians behavior in crowded scenes. People walking in a group tend to keep their formations for a short time. This provides valuable information to any tracking algorithms. However, the biggest challenge is: \emph{How to incorporate group information in a tracking framework?} To utilize group information one needs to incorporate pairwise information in the tracking formulation. In our case, this is done by simply adding another quadratic term in our objective function that captures the group formations. 

\begin{figure}
	\centering
	\includegraphics[width=1\linewidth]{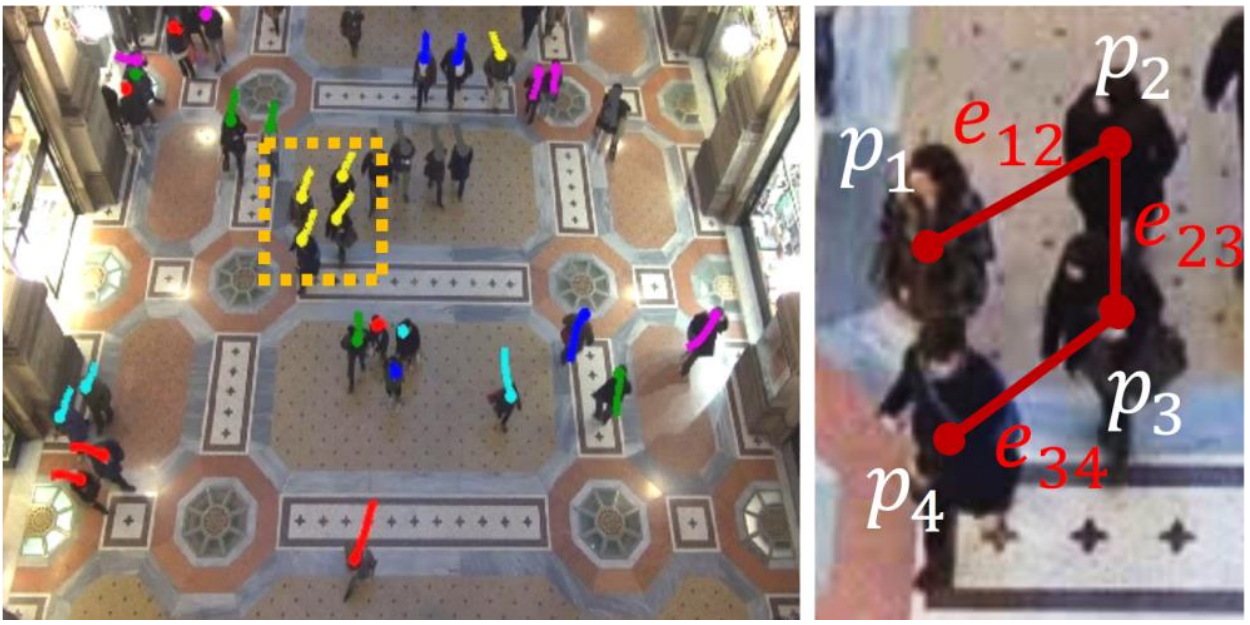}
	\caption{An example of our minimum spanning tree group structure model. The figure on the left shows the groups found in frame $638$ of Galleria1 sequence (nearby tracks with similar color correspond to one group). The figure on the right illustrates the model created for the group inside the yellow box. $p_i$ identifies the location of the target and $e_ij$ determines the relative distance between the two targets $i$ and $j$. }
	\label{fig:group}
\end{figure}

Let $G_i=\{ p_1, p_2, ..., p_m\}$ be the $i^{th}$ group which contains $m$ targets, where $p_j$ defines the location of the $j^{th}$ target. We adopt a minimum spanning tree pictorial structure model to represent each group. The main advantage of having a tree model instead of considering all the pairwise relationships is bifold. First, considering fewer pairwise relationships, helps reducing the computational complexity. Second, having fewer constraints is less restrictive and better allows small changes in target formations, which is likely to happen in practice. The parameters of our pictorial structure model are also learned online during tracking, and do not involve large training data like most previous works. The grouping information in our formulation is encoded by the matrix $\textbf{C}_g$, where $\mathbf{C}_g=I-D^{-1/2} \Gamma  D^{1/2}$ is the normalized Laplacian matrix \cite{graphcut}. Here $D$ is the diagonal matrix composed of row sums of similarity matrix $\Gamma$. $\Gamma \in \mathbb{R}^{(l) \times l}$ is the similarity matrix that encodes the groping information and encourages the selection of candidate locations that keep the formation of targets within each group. Each entries of $\Gamma$ is obtained using the equation below:

\begin{equation}
\Gamma_{ij}=exp(\frac{-( \left \| p_i-p_j \right \|_2 - e_{ij})}{2\sigma^2}),
\label{eq:groupSimilarity}
\end{equation}

\noindent where $e_{ij}$ is the distance between target $i$ and target $j$ in our tree model for that group. An example of our tree model for one group is shown in Figure \ref{fig:group}. We update the group information every $\tau$ frames ($\tau = 10$ in our experiments), thus the values of $e_{ij}$ are updated every $\tau$ frames.  We found that it is important to update the groups frequently, because targets within the same group are likely to change their relative distance. 

\section{Optimization}
\label{sec:FW}

Since $\mathbf{C}_{sp}$ and $\mathbf{C}_{g}$ are positive semi definite, one can use publicly available QP software such as ILOG CPLEX to find the solution to Eq. \ref{eq:cost}. However, we observed that the solver becomes extremely slow as the number of targets exceeds five, thus it is not feasible to solve the BQP directly. One option is to relax the discrete non convex binary constraint in Eq. \ref{eq:const2} to its convex hull. The barrier optimization techniques used in commercial softwares such as CPLEX has complexity of $O(N^3)$, which makes it inefficient when solving for hundreds of targets. On the other hand, the structure of our problem allows one to solve the linearized version of Eq. \ref{eq:cost} very efficiently using projected gradient descend methods. This property opens the room to the powerful Frank Wolfe optimization technique that has been revisited recently. We adopt the most recent version of Frank Wolfe algorithm that takes advantage of the SWAP steps to speed up the optimization. We show that using Frank-Wolfe with SWAP steps one can find the solution efficiently for hundreds of targets. We also compare the run-time of our optimization with CPLEX and other variants of Frank-Wolfe algorithm and show that we can find the solution faster. 

\subsection{Frank Wolfe Optimization}

\begin{algorithm}
\small
	\DontPrintSemicolon
	\KwData{$\mathbf{x}_0 \in {D}$.}
	\KwResult{$\mathbf{z}$}
	\KwResult{Initialization: $k=0$, ${\mathbf{z}}={\mathbf{x}}_0$}
	\For{$k=0,1...K$}{
		Compute ${\mathbf{s}}_k\leftarrow \underset {{\mathbf{s}} \in D} {\text{argmin}}<{\mathbf{s}},\bigtriangledown f({\mathbf{x}}_k)>,$ \\
		$d_{FW}={\mathbf{s}}_k-{\mathbf{x}}_k$\\
		Line Search : $\lambda_{FW} = \underset {\lambda \in [0,1]} {\text{argmax}} f({\mathbf{x}}_k+\lambda(d_{FW}))$\\
		Update : ${\mathbf{x}}_{k+1}=(1-\lambda_{FW}){\mathbf{x}}_k+(\lambda_{FW}){\mathbf{s}_k}$
	}
	Perform the rounding $\mathbf{z} \leftarrow rounding({\mathbf{x}}_K)$ \\
	return $\mathbf{z}$	
	\caption{Frank Wolfe}
	\label{alg:FW}
\end{algorithm}

Given our convex quadratic function $f(\mathbf{x})$ in Eq. \ref{eq:cost} and a set of convex constraints $D$, Frank Wolfe algorithm finds a solution to this problem by solving the iterative optimization given in Algorithm. \ref{alg:FW}. At every iteration it solves the linearized version of the objective function, $g(\mathbf{x})$. The minimizer is then used to find the descent direction after performing a line search. In our problem, the linearized version of our objective function is given by \footnote{please note that for simplicity, we removed the index $k$ in Equations \ref{eq:linearizedFunction}, \ref{eq:lineSearch} and \ref{eq:step} , which shows the $k^{th}$ iteration}:

\begin{equation}
\begin{split}
g(\mathbf{x}) &= <{\mathbf{s}},\bigtriangledown f({\mathbf{x}})>  \\
&= {\mathbf{s}}(\mathbf{C}_{sp}{\mathbf{x}}+ \mathbf{C}_{g}{\mathbf{x}} + \mathbf{c}_a + \mathbf{c}_m + \mathbf{c}_{nm}),
\end{split}
\label{eq:linearizedFunction}
\end{equation}

\noindent subject to the set of convex constraints defined by $D$. This could be solved efficiently using any projected gradient descend method. One can define the step size in Algorithm. \ref{alg:FW} by $\lambda_{FW}=2/(2+k)$, where $k$ is the current iterate. However, the exact step size is found by solving the line-search problem given below:

\begin{equation}
\lambda_{FW} = \underset {\lambda \in [0,1]} {\text{argmax}} f({\mathbf{x}}+\lambda({\mathbf{s}}-{\mathbf{x}})).\\
\label{eq:lineSearch}
\end{equation}

The optimization problem in Equation. \ref{eq:lineSearch} has a closed form solution that can be obtained by setting the derivative of Equation \ref{eq:lineSearch} with respect to $\lambda$ equal to zero, and replacing the function $f$ with its definition in Equation \ref{eq:cost}. This will lead to the following closed form solution:

\begin{equation}
\begin{split}
\frac{\partial }{\partial \lambda}f({\mathbf{x}}+\lambda ({\mathbf{s}}-{\mathbf{x}}))&=\bigtriangledown f({\mathbf{x}}+\lambda({\mathbf{s}}-{\mathbf{x}}))^T({\mathbf{s}}-{\mathbf{x}})=0, \\
\lambda_{FW}&=\frac{\bigtriangledown f({\mathbf{x}})^T({\mathbf{s}}-{\mathbf{x}})}{({\mathbf{s}}-{\mathbf{x}})(\mathbf{C}_{sp}+\mathbf{C}_{g})({\mathbf{s}}-{\mathbf{x}})}.
\end{split}
\label{eq:step}
\end{equation}

One should note that the solution in Equation \ref{eq:step} will not add much overhead to the computation. Because most of the terms have been already computed in other steps. The only part in Algorithm \ref{alg:FW} that is left unexplained is the rounding. The solution found at the end of FW optimization does not satisfy the discrete constraint of Eq. \ref{eq:const2}, thus requires rounding. In order to find the best binary solution one needs to solve the following optimization. 

\begin{equation}
 \underset {{\mathbf{y}} \in D} {\text{argmin}} \left \| {\mathbf{y}}-{\mathbf{x}}_K \right \|_2, 
\label{eq:rounding}
\end{equation}

\noindent where $\textbf{x}_K$ is the solution found by FW at the end of the optimization at iteration $K$ and $\textbf{y}$ is the final rounded solution. Due to the structure of our problem, the optimization in \ref{eq:rounding} reduces to solving the above optimization for each target separately, which is equivalent to taking the $\text{argmin}$ of the $\mathbf{x}$ for those candidates of each target. 

\subsection{Frank Wolfe with SWAP Steps}

The convergence rate of original Frank-Wolfe algorithm is shown to slack near the optimal solution. This makes the original FW method intractable for large scale problems. Instead of using the original FW, we use an accelerated version of Modified-Frank-Wolfe algorithm that takes advantage of a trick called SWAP steps to speed up the optimization. The full optimization procedure is given in Algorithm \ref{alg:FWSWAP}. The idea is that at each iteration we find the descend vertex, $\mathbf{x}_k$ as well as the ascend vertex, $\mathbf{y}_k$, over the face spanned by current solutions. Beside the FW step of $d_{fw}=\mathbf{x}_k-\mathbf{z}$, we consider the SWAP step defined as $\mathbf{x}_k-\mathbf{y}_k$. The SWAP could be considered as a step that moves the current solution in the away direction and at the same time in the direction of the toward step. This is shown in the equation below. 

\begin{figure}
	\centering
	\includegraphics[width=1\linewidth]{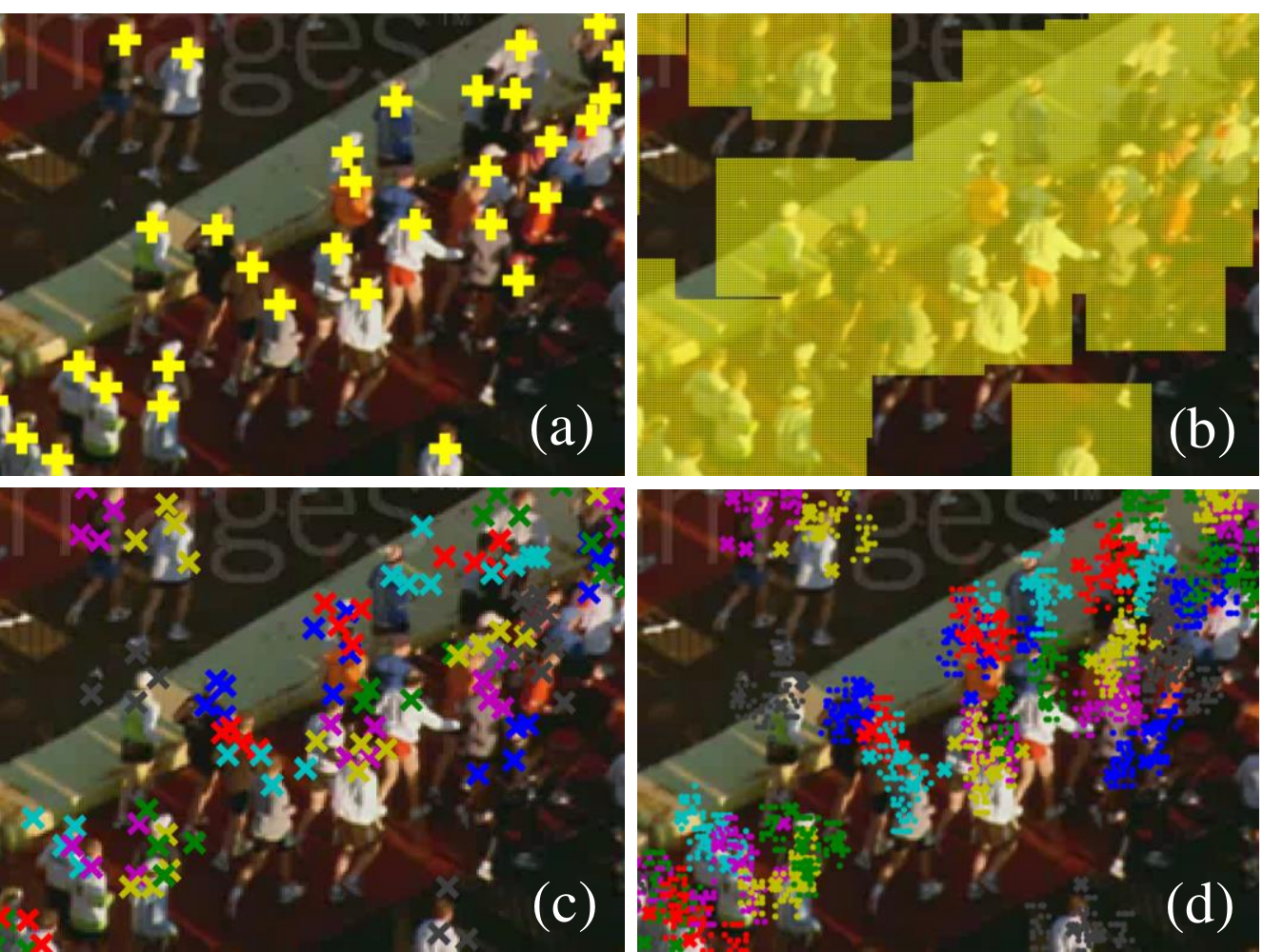}
	\caption{A comparison of our speed-up sampling technique vs dense sampling of candidate locations. (a) shows the targets to be tracked. (b) illustrates the candidates sampled densely in the neighborhood of each target. (c) shows the selected extrema locations (each candidate is shown with a cross). (d) visualizes the final candidates that survived using the proposed speed-up technique (each candidate is shown with a small dot). }
	\label{fig:SpeedUp}
\end{figure}

\begin{equation}
\begin{split}
\overbrace{\mathbf{x}_k+\lambda(\mathbf{s}_k-\mathbf{y}_k)}^{SWAP \ step} &= \overbrace{\frac{1}{2}(\mathbf{x}_k+\lambda(\mathbf{s}_k-\mathbf{x}_k))}^{toward \ step}\\ &+ \overbrace{\frac{1}{2}(\mathbf{x}_k+\lambda(\mathbf{x}_k-\mathbf{y}_k))}^{away \ step}. 
\end{split}
\label{eq:swapStep}
\end{equation}

\begin{algorithm}
\small
\DontPrintSemicolon
\KwData{$\mathbf{x}_0 \in \textsl{D}$, $\varepsilon >0$.}
\KwResult{$\mathbf{z}$}
\KwResult{Initialization: $k=0$, $\mathbf{z}=\mathbf{x}_0$, $S_0=\{ \mathbf{x}_0 \}$, $max\_it$}
\While{$duality\_gap(\mathbf{z})>\varepsilon$ and $k<max\_it$}{
$k\leftarrow k+1$\\
(descent direction) $\mathbf{s}_k \leftarrow \underset {\mathbf{s} \in D} {\text{argmin}}<\mathbf{s},\bigtriangledown f(\mathbf{x}_k)>$ \\
(ascent direction)$\mathbf{y}_k \leftarrow \underset {\mathbf{y} \in S_{k-1}} {\text{argmax}}<\mathbf{y},\bigtriangledown f(\mathbf{x}_k)>$ \\
Line Search : $\lambda_{fw} = \underset {\lambda \in [0,1]} {\text{argmax}} f(\mathbf{x}_k+\lambda(\mathbf{s}_k-\mathbf{x_k}))$\\
Line Search : $\lambda_{swap} = \underset {\lambda \in [0,1]} {\text{argmax}} f(\mathbf{x}_k+\lambda(\mathbf{s}_k-\mathbf{y}_k))$\\
Compute $\delta_{fw}=f(\mathbf{x}_k+\lambda_{fw}(\mathbf{s}_k-\mathbf{x}_k))-f(\mathbf{x}_k)$ (Improvement of fw step) \\
Compute $\delta_{swap}=f(\mathbf{x}_k+\lambda_{swap}(\mathbf{s}_k-\mathbf{y}_k))-f(\mathbf{x}_k)$ (Improvement of SWAP step) \\
Compute $\delta_k = max(\delta_{swap},\delta_{fw})$\\
      \If{$ \delta_k = \delta_{swap}$}{
      Clip the line serach parameter, $\lambda_{swap*}=max(\lambda_{swap},\alpha_k(\mathbf{y}_k))$ \\
      if $\lambda_{swap*}=\alpha_k(\mathbf{y}_k)$ mark it as SWAP-drop step \\
      if $\lambda_{swap*}=\lambda_{swap}$ mark it as SWAP-add step \\      
        Perform the SWAP step $\mathbf{x}_{k+1} = \mathbf{x}_{k} + \lambda_{swap*}(\mathbf{s}_k-\mathbf{y}_k)$\\
        }
        \Else{
      Perform the FW step \\       
      $\mathbf{x}_{k+1} = \mathbf{x}_{k} + \lambda_{fw*}(\mathbf{s}_k-\mathbf{x}_k)$
        }
      }
    Perform the rounding $\mathbf{z} \leftarrow rounding(\mathbf{x}_K)$ \\
    return $\mathbf{z}$

\caption{Frank Wolfe with SWAP Steps}
\label{alg:FWSWAP}
\end{algorithm}

Once we define the toward and SWAP steps, we find the improvement update using each step. The one that gives the best improvement is selected to perform the move in the current iteration. In order to pick the best improvement one needs to perform two line-searches compared to one line-search step that was used in previous accelerated versions of Frank-Wolfe \cite{JoulinVidLocECCV14}. However, since the estimation of the objective function at each iteration is more accurate, it requires less iterations to converge. Additionally, the optimal value of line-segment problem is found analytically and does not require much computations. The computation of $\delta_{fw}$ and $\delta_{swap}$ involves terms already computed in the line-search and therefore does not add any additional overload.  We present several experiments in Section \ref{sec:experiments} to validate the discussions above. Clipping the line search is to ensure the solution remains in the convex set  $\textsl{D}$. SWAP-add/drop step is used to update the active set $S$ \cite{simon2015}.

\begin{table*}[ht!]
	\caption{Quantitative results of our method in terms of \textbf{Tracking Accuracy} when pixel threshold is set to $15$. We compared our method with six competitors on nine sequences of \cite{IdreesIVC14}. On average we improve the best previous tracker by $3.1\%$ in nine sequences.}
	\label{tab:quantitative-overall}
	\centering
	\begin{tabular}{ c | c | c | c | c | c | c | c | c | c }
		\hline
		& Seq1 & Seq2 & Seq3 & Seq 4 & Seq 5 & Seq 6 & Seq 7 & Seq 8 & Seq 9 \\ \hline \hline
		\#Frames  & 840 & 134 & 144 & 492 & 464 & 133 & 494 & 126 & 249 \\ 
		\#People  & 152 & 235 & 175 & 747 & 171 & 600 & 73 & 58 & 57\\  \hline
		NCC & 49\% & 85\% & 58\% & 52\% & 33\% & 52\% & 50\% & 86 & 33\% \\
		MS & 19\% & 67\% & 16\% & 8\% & 7\% & 36\% & 28\% & 43\% & 10\%\\
		MSBP & 57\% & 97\% & 71\% & 69\% & 51\% & 81\% & 68\% & 94\% & 40\%\\
		FF     & 74\% & 99\% & 83\% & 88\% & 66\% & 90\% & 68\% & 93\% & 47\% \\
		CTM & 76\% & 100\% & 88\% & 92\% & 72\% & 94\% & 65\% & 94\% & 66\%\\
		NMC & 80\% & 100\% & 92\% & 94\% & 77\% & 94\% & 67\% & 92\% & 63\%\\
		\definecolor{LightCyan}{rgb}{0.88,1,1}
		\textbf{Proposed} & \textbf{86\%} & \textbf{99\%} & \textbf{96\%} & \textbf{97\%} & \textbf{78\%} & \textbf{96\%} & 67\% & 90\% & 78\% \\ \hline		
	\end{tabular}
\end{table*}

\section{Speed-UP}
\label{sec:speedUP}

Although Frank-Wolfe algorithm speeds up the optimization significantly, we noticed that there is a room for even further speed up. This is important specially when the number of targets in the scene reaches to a few hundreds. The main motivation behind this is that, a lot of candidate locations can be removed which leads to reduction in the number of variables in the optimization (An example is shown in Figure \ref{fig:SpeedUp}(b,d)). One naive way of removing the undesired candidate locations is thresholding the confidence values of our detector  ($\mathbf{c}_a$)  or thresholding the confidence value of all three linear terms ($\mathbf{c}_a + \mathbf{c}_m + \mathbf{c}_{nm}$). This is similar to what a pre-trained object detector does.  However, this may results in removing useful candidates that represent the targets and are assigned low scores due to pose changes or occlusion. Moreover, keeping only the samples with high confidence values in the linear terms (Figure \ref{fig:SpeedUp}(c)) in Equation \ref{eq:cost}, limits the effect of our quadratic terms that capture the spatial proximity as well as grouping. 

Instead, we incorporate a better way of sampling candidate locations, which does not drop the accuracy and at the same time is capable of showing the effect of each term in our optimization. Our sampling starts with first selecting the top $m$ extrema points in the probability map obtained from the linear terms in Equation \ref{eq:cost} ($m$ is set to three). An example is shown in Figure \ref{fig:SpeedUp}(c). In order not to limit ourselves to the high confidence locations found by $\mathbf{c}_a + \mathbf{c}_m + \mathbf{c}_{nm}$, we further sample an extra $10$ candidates in a small neighborhood, ($6 \times 6$) of each extrema point (Figure \ref{fig:SpeedUp}(d)). The latter step will allow the quadratic terms to make the necessary changes to the target locations in order to improve the optimization cost.  We observed in our experiments that, if only the extrema points are selected, the average performance for the 9 sequences is $1.5\%$ lower than the ones reported in Table \ref{tab:quantitative-overall}. However, when we use our sampling technique, the performance is only $0.2\%$ lower compare to when all the candidate locations are used. 

\begin{figure}[ht!]
	\centering
	\includegraphics[width=0.7\linewidth]{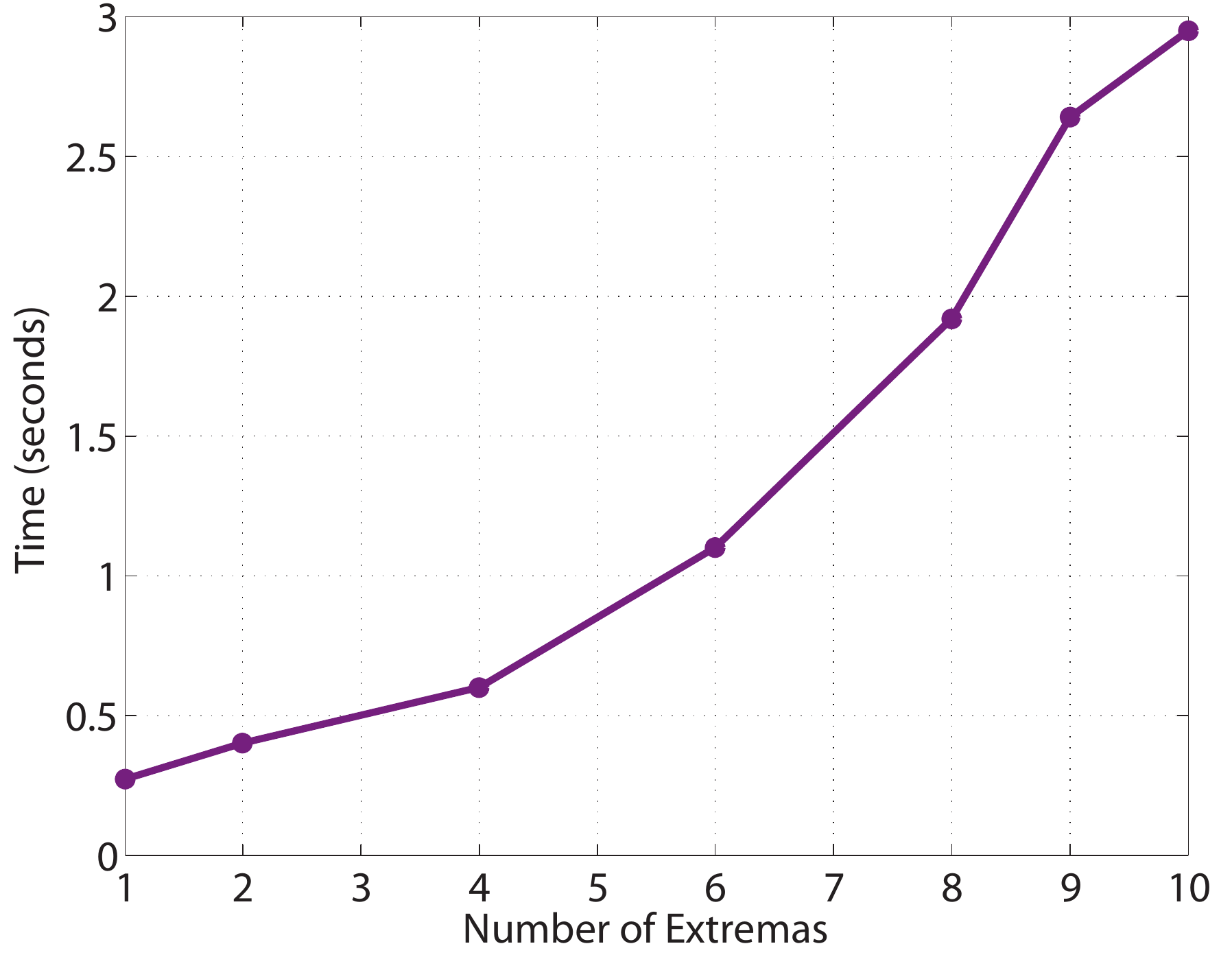}
	\caption{This figure illustrates the run-time vs the number of extremas ($m$).}
	\label{fig:accVsSample}
\end{figure}

An example of our sampling technique is shown in Figure \ref{fig:SpeedUp}. This procedure reduces the number of candidate more than an order of magnitude which results in significant speed up. Furthermore, in Figure \ref{fig:accVsSample} we show the effect of $m$ in the speed-up. As can be seen when the number of candidates are dropped by an order of magnitude, we get almost six times speed-up in the optimization. As also mentioned earlier, the performance almost remains the same when we set $m=3$.

\section{Experiments}
\label{sec:experiments}

We perform exhaustive experiments on nine high density crowd sequences of \cite{IdreesIVC14} and two new sequences with medium crowd density. These sequences include different scenarios and challenges. All sequences are taken using cameras facing down. The parameters $\zeta$ and $\eta$ are set to $0.3$ and $0.2$, which we found through cross validation. The search region for each target is set to twice the size of the target for all sequences.   

\begin{figure*}[ht!]
	\centering
	\includegraphics[width=0.85\linewidth]{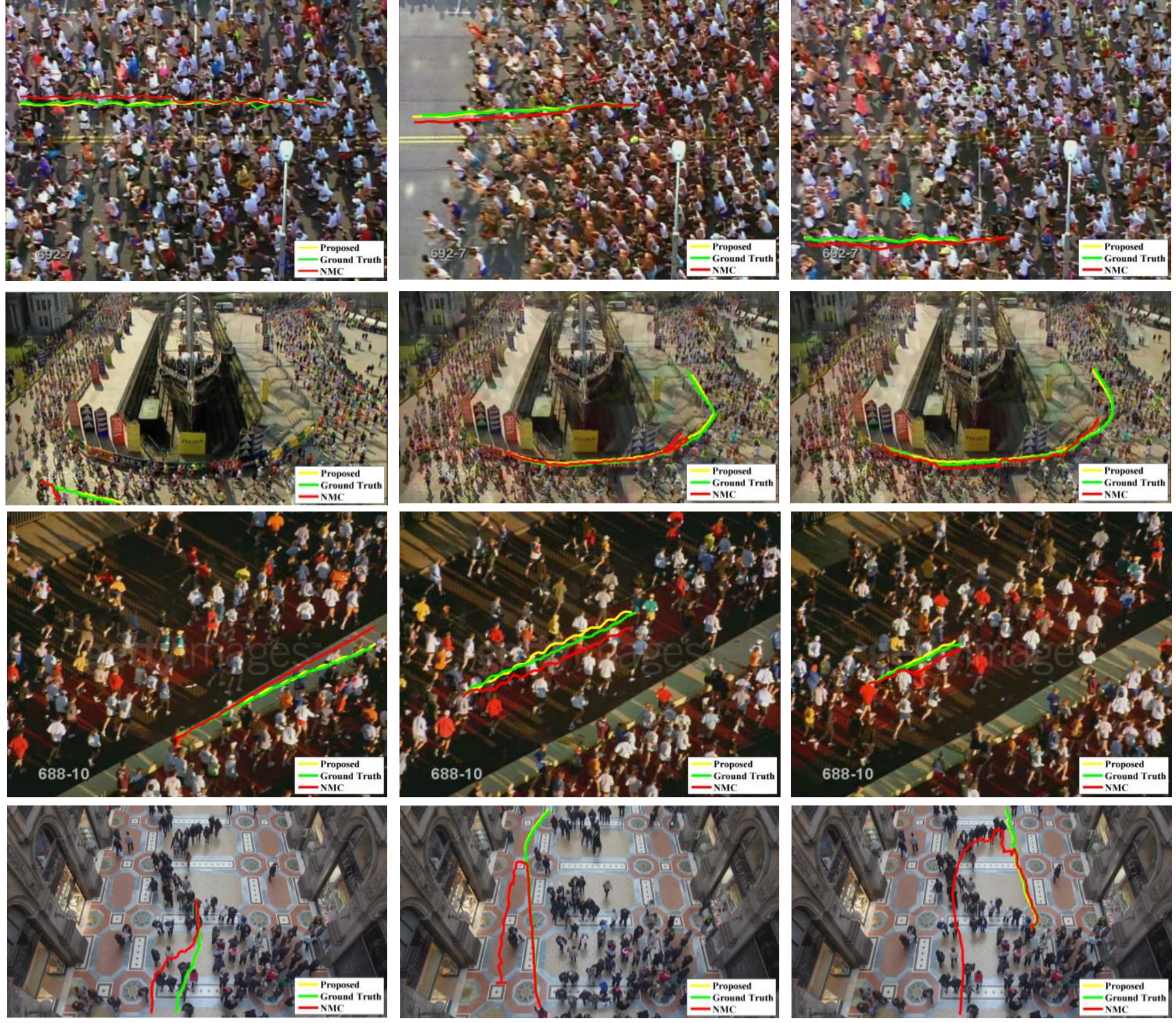}
	\caption{This figure shows quantitative results of our method for a few challenging targets. We also compare our results with the one in \cite{IdreesIVC14}. For some sequences the proposed method gives tracks very close to the ground truth tracks. Since green is superimposed on yellow, due to that tracks in yellow may not be that visible. }
	\label{fig:qualitative-results}
\end{figure*}

\textbf{High-density Crowd Sequences: }First frames of the high density crowd sequences of \cite{IdreesIVC14} are shown in Figure \ref{fig:results}. Seq-3, seq-4, seq-5 and seq-6 show marathon events where targets with similar appearance run close to each other. Seq-1 shows daily commute of crowds. Targets look very similar to their background and they often get confused with the background. Seq-2 is taken from Hajj event, seq-7 shows the crowd in a train station. Seq-8 is taken from airport lobby and seq-9 contains people crossing a street. This dataset contains structured and unstructured crowd sequences with lots of anomalies even in structured crowd sequences. The crowd density in each video is different. Some statistics of the dataset are shown in Table. \ref{tab:quantitative-overall}. The number of individuals annotated in each sequences ranges from $57$ to $747$. 

\textbf{Medium-density Crowd Sequences: }
We annotated two new sequences, in addition to the high density crowd sequences, to show effectiveness of our method for medium density crowd sequences as well. We have named them, Galleria1 and Galleria2.  These two sequences are recorded from Galleria mall in Milan and have been used in the vision community for other tasks such as group detection \cite{RitaPAMI2015}. We annotated the first $2000$ frames of each sequence and included them in our evaluation. With the permission from the group which published the sequences \cite{MilanoGroup}, we plan to release the videos along with their annotations upon the acceptance of the paper. Galleria1 contains 200 targets and Galleria2 contains 215 targets. Some example frames of these two sequences along with qualitative results are shown in Figure \ref{fig:qualitative-results}. 

\subsection{Overall Performance}

We compare our method with previous methods designed to track individuals in high-density crowds. Below we provide a summary of each approach: 

\begin{itemize}

\item Floor Fields method of Ali and Shah (FF) \cite{AliECCV2008} : This method is based on the assumption that all targets follow global crowd behavior at every location in the scene. The prior information they learn, called floor fields, restrict the motion of each individual in a scene severely. 

\item Correlated Topic Model (CTM) \cite{RodriguezCVPR2009}:  CTM tracker utilizes a Correlated Topic Model to model crowd behavior at each location. Their model does not have the assumption of \cite{AliECCV2008} in which targets are restricted to take only one direction at each location. But still it needs to learn dynamic model of the scene given some training data. In their construction, words correspond to low level quantized motion features and topics correspond to crowd behaviors

\item Mean-shift Belief Propagation (MSBP) \cite{MSBP}: MSBP tracker models the contextual relationship between the target in an MRF framework. The mean-shift belief propagation technique was used for the optimization. 

\item Prominence Neighborhood Motion Concurrence (NMC) \cite{IdreesIVC14}: The PNMC tracker utilized a template based tracker at its core. The targets are tracked individually in an ordered fashion employing information from the neighborhood and confidence from the template based tracker.  

\end{itemize}

We used the manual annotation of initial target locations provided with the dataset. The template size is the same as the one used in \cite{IdreesIVC14}. 

\begin{table}
	\centering
		\caption{Quantitative Comparison, in terms of tracking accuracy, of our method with the tracker of \cite{IdreesIVC14} when pixel threshold is set to 15 on two new sequences of Galleria1 and Galleria2. On average we improve NMC tracker of \cite{IdreesIVC14} by $4.5\%$ on these two sequences.}.
	\label{tab:quantitative-overall-milano}
	\begin{tabular}{ c | c | c }
		\hline
		& Galleria1 & Galleria2  \\ \hline \hline
		\#Frames  & 2000 & 2000  \\ 
		\#People  & 200 & 215 \\  \hline
		PNMC & 86\% & 88\% \\
		\textbf{Proposed} & \textbf{92\%} & \textbf{91\%} \\ \hline		
	\end{tabular}
\end{table}

\subsection{Quantitative Results}
\begin{figure*}[ht!]
	\centering
	\includegraphics[width=1\linewidth]{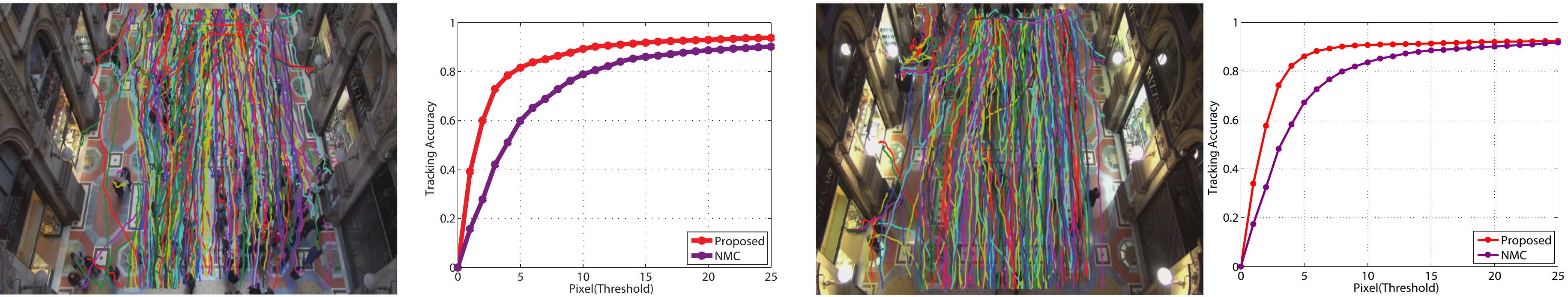}
	\caption{This figure shows tracks (show on the left) and quantitative results of our method. We compare our method with NMC \cite{IdreesIVC14} which achieves the best results on the $9$ high-density crowd sequences.}
	\label{fig:quantitative-milano}
\end{figure*}

\begin{figure*}[ht!]
	\centering
	\includegraphics[width=1\linewidth]{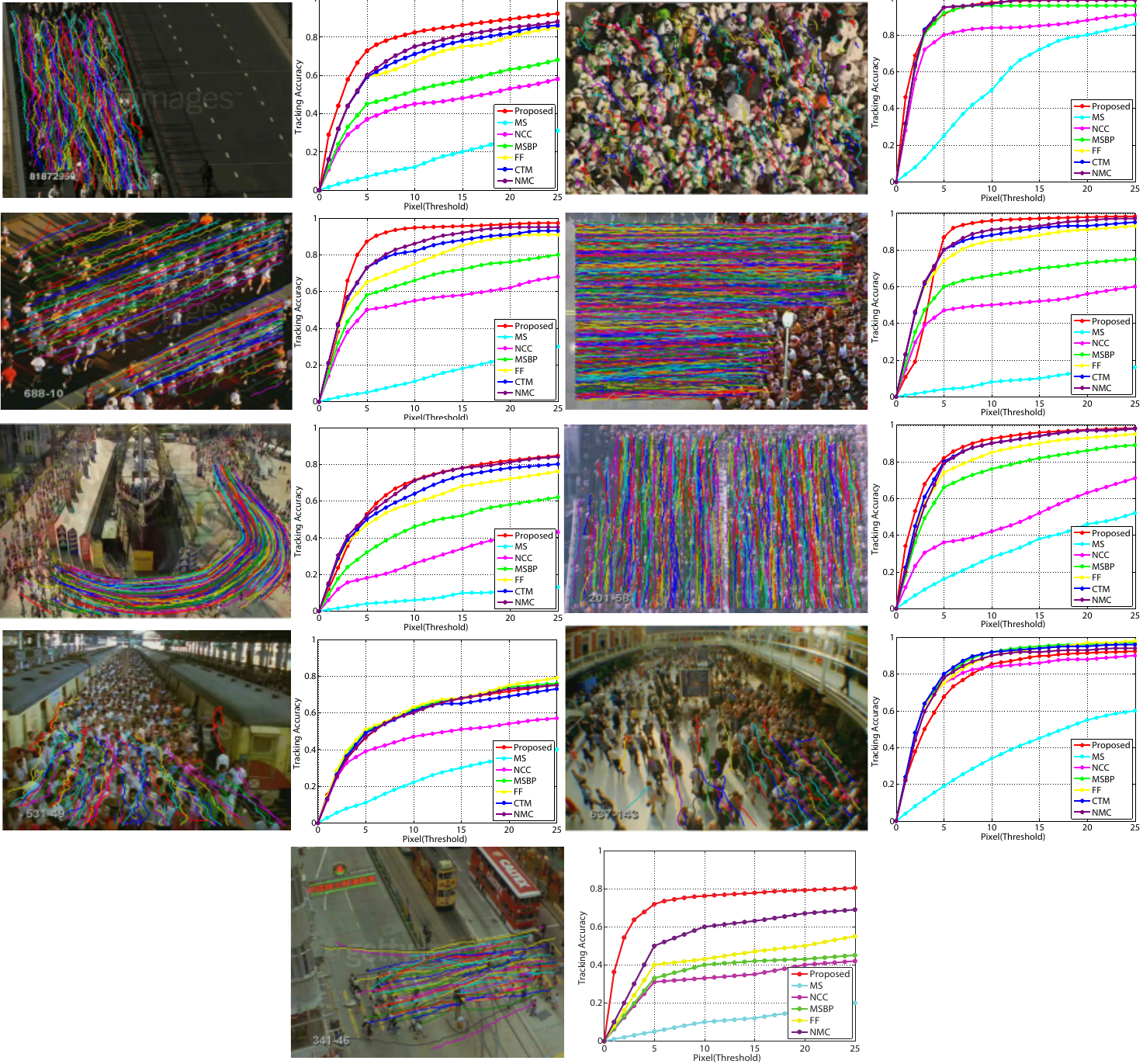}
	\caption{This figure shows tracks (show on the left) and quantitative results of our method with competitive approaches of FF\cite{AliECCV2008}, CTM\cite{RodriguezCVPR2009}, MSBP \cite{MSBP}, PNMC \cite{IdreesIVC14} and MS\cite{MS}. For each sequence we show the qualitative results of all tracks and on the right we show the quantitative comparison. Each plot shows the tracking accuracy vs pixel error.  }
	\label{fig:results}
\end{figure*}

\begin{figure*}[ht!]
	\centering
	\includegraphics[width=0.9\linewidth]{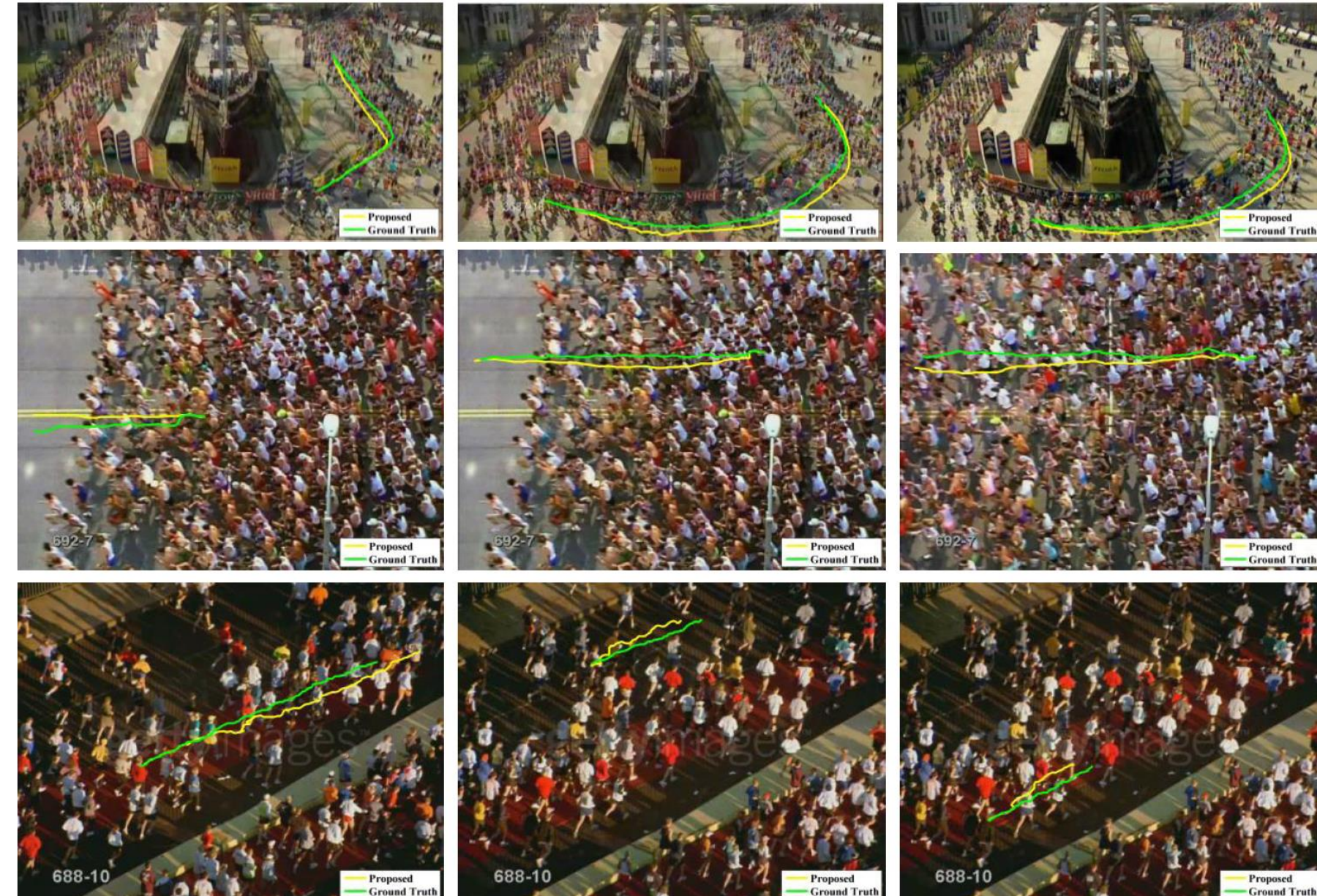}
	\caption{This figure shows the failure cases of our method in three different sequences. The failure cases mostly belong to the targets with inaccurate initialization which leads to poor appearance information. The drift usually happens in the first few frames.}
	\label{fig:failure}
\end{figure*}

We followed the same metrics as the one in \cite{IdreesIVC14} to quantitatively compare our method with others. The tracking accuracy for different pixel-error threshold is shown in Figure \ref{fig:results} for the $9$ high-density crowd sequences of \cite{IdreesIVC14}. The results on Galleria1 and Galleria2 sequences are also shown in Figure \ref{fig:quantitative-milano}. Similar to \cite{IdreesIVC14} we also provide the accuracy for when the pixel threshold is set to $15$ in Tables \ref{tab:quantitative-overall} and \ref{tab:quantitative-overall-milano}. Our method outperforms the existing approaches in most sequences. The performance increase in sequence $9$ is worth a special mention here. In this sequences people are walking in the opposite directions, which is the scenario that \cite{AliECCV2008} is not designed to handle. The heuristic such as instantaneous flow proposed in\cite{IdreesIVC14} are not suitable for tracking in these scenarios. In Figure \ref{fig:failure}, we show some of the failure cases of our method. 

\subsection{Contribution of the Proposed Terms}

In order to show the effectiveness of the proposed terms in the objective function, we conducted an experiment with different setups. The detailed results for each sequence are shown in Table. \ref{tab:quantitative}. We tried a combination of different terms. Our baseline (B(NCC)) is just a template-based tracker without the other terms in Eq. \ref{eq:cost}. Comparing the baseline of \cite{IdreesIVC14} that also uses a NCC based tracker with the equivalent of ours (B+Mo(NCC)), one can see that formulating the problem as multi-target tracking with joint optimization of target tracks can improve the performance significantly. Our baseline is $30\%$ higher that the one of \cite{IdreesIVC14}.

\begin{table*}
	\caption{Quantitative comparison of \textbf{Tracking Accuracy} using different terms in cost function of Equation. \ref{eq:cost}.B is the baseline. Mo is the motion term ($c_m$), SP is the spatial proximity term($c_{sp}$), NMo is neighborhood motion cost ($c_{nm}$). NCC represent the template based tracker based on Normalized Cross Correlation and KCF represents the discriminative model based on kernalized correlation filters.}
	\label{tab:quantitative}
	\centering
	\begin{tabular}{ c | c | c | c | c | c | c | c | c | c }
		\hline
		& Seq1    & Seq2 & Seq3 & Seq 4 & Seq 5 & Seq 6 & Seq 7 & Seq 8 & Seq 9 \\ \hline \hline
		\#Frames  & 840 & 134 & 144 & 492 & 464 & 133 & 494 & 126 & 249 \\ 
		\#People  & 152 & 235 & 175 & 747 & 171 & 600 & 73 & 58 & 57\\  \hline
		B(NCC)          & 78.13\% & 98.54\% & 85.54\% & 91.18\% & 65.34\% & 93.49\% & 66.81\% & 88.76   & 63.72\% \\
		B+Mo(NCC)       & 76.9\%  & 98.83\% & 87.64\% & 92.68\% & 72.18\% & 92.95\% & 73.80\% & 91.74\% & 64.08\%\\
		B+Mo+SP(NCC)    & 77.29\% & 99.03\% & 93.70\% & 92.28\% & 76.08\% & 94.28\% & \textbf{74.37}\% & \textbf{93.00}\% & 67.73\%\\
		B+Mo+Gr(NCC)    & 79.74\% & 99.03\% & 93.13\% & 93.73\% & 77.08\% & 94.17\% & 72.92\% & 93.00\% & 64.07\%\\
		B+Mo+SP+Gr+NMo(NCC)& 80.32\% & 99.03\% & 93.40\% & 95.18\% & 76.67\% & 94.65\% & 74.0\%  & 92.13\% & 69.80\% \\
		B+Mo+SP+Gr+NMo(KCF)& \textbf{86.08\%} & \textbf{98.62\%} & \textbf{96.41\%} & \textbf{96.84\%} & \textbf{77.73\%} & \textbf{95.75\%} & 67.22\% & 90.35\% & \textbf{77.88}\% \\ \hline
	\end{tabular}
\end{table*}

We next add different components of the objective function in \ref{eq:cost} to the baseline tracker one by one, in order to evaluate their potency. $B+Mo$ is our baseline tracker where we add the linear motion constraint to it. $B+Mo+SP$ and $B+Mo+Gr$ are the same as $B+Mo$ with an additional spatial proximity constraint and group constraint respectively. We later add the neighborhood motion term ($B+Mo+Sp+NMo$) to evaluate its effectiveness, and finally we replace the generative template based tracker with our discriminative model ($B+Mo+Ov+NMo(disc)$)and show that it further improves the overall performance. 
From the results in Table \ref{tab:quantitative}, we observe that the improvement of different terms depends on different factors in the scene including the density of crowd. For example in sequence $7$ and $8$ the performance slightly decreases when adding the neighborhood motion term. This is mostly due to heterogeneity of target movement in those sequences. Spatial proximity and grouping terms in all the sequences improve the performance which show their effectiveness. One should note that people rarely form groups in extremely crowded scenes. However, our observation shows that enforcing the consistency in formation of the coherently moving targets that are close to each other will help improving the results.

\begin{figure}[ht!]
	\centering
	\includegraphics[width=0.65\linewidth]{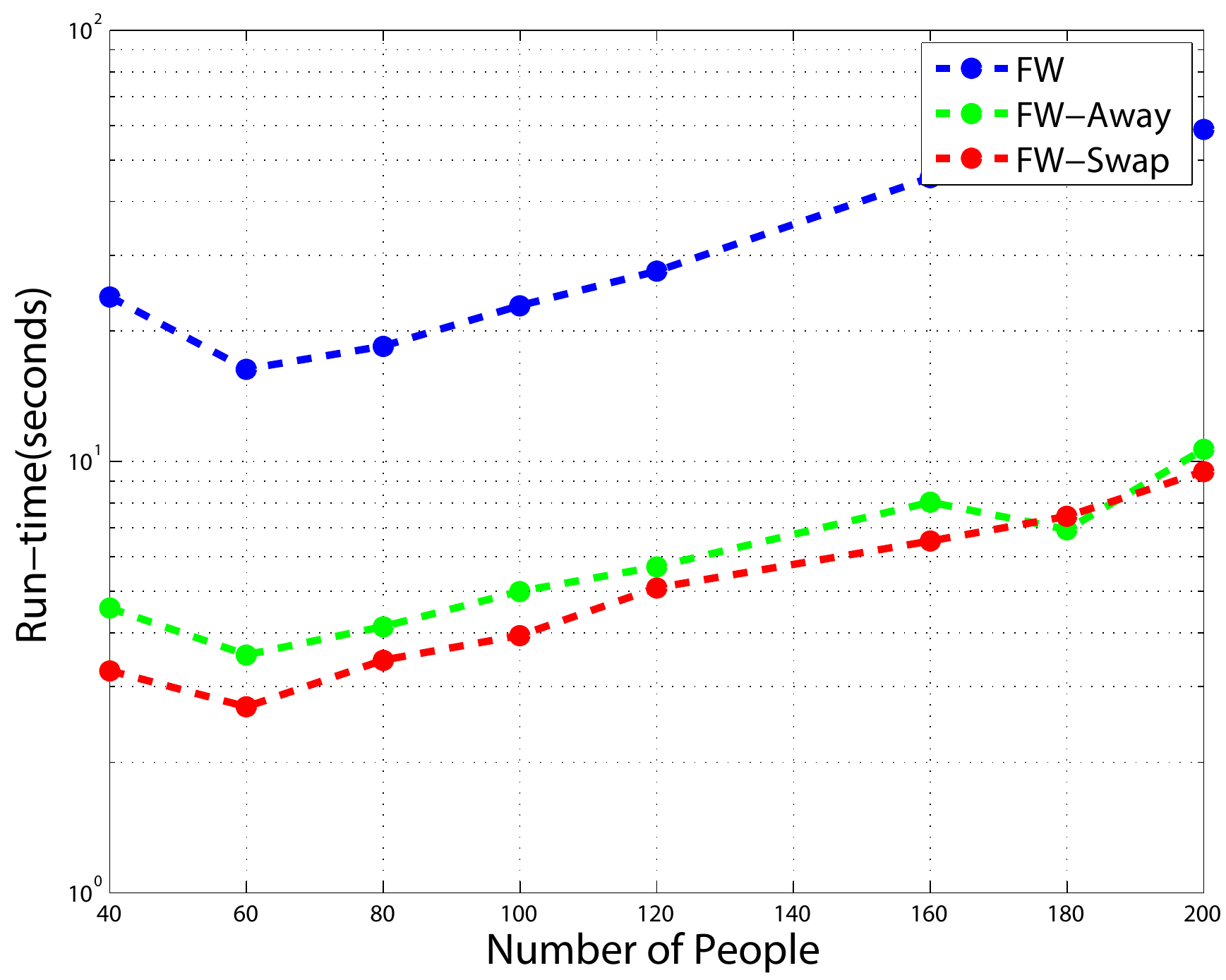}
	\caption{The figure shows the run-time comparison of Frank-Wolfe(FW), Frank-Wolfe with Away steps (FW-Away) and Frank-Wolfe with Swap steps (FW-Swap) on one of our sequences. It is clear that the Away step technique improves the run-time of FW significantly. However, using the Swap step we can further speed-up the optimization.}
	\label{fig:runTimeFW}
\end{figure}

\begin{figure}[ht!]
	\centering
	\includegraphics[width=0.65\linewidth]{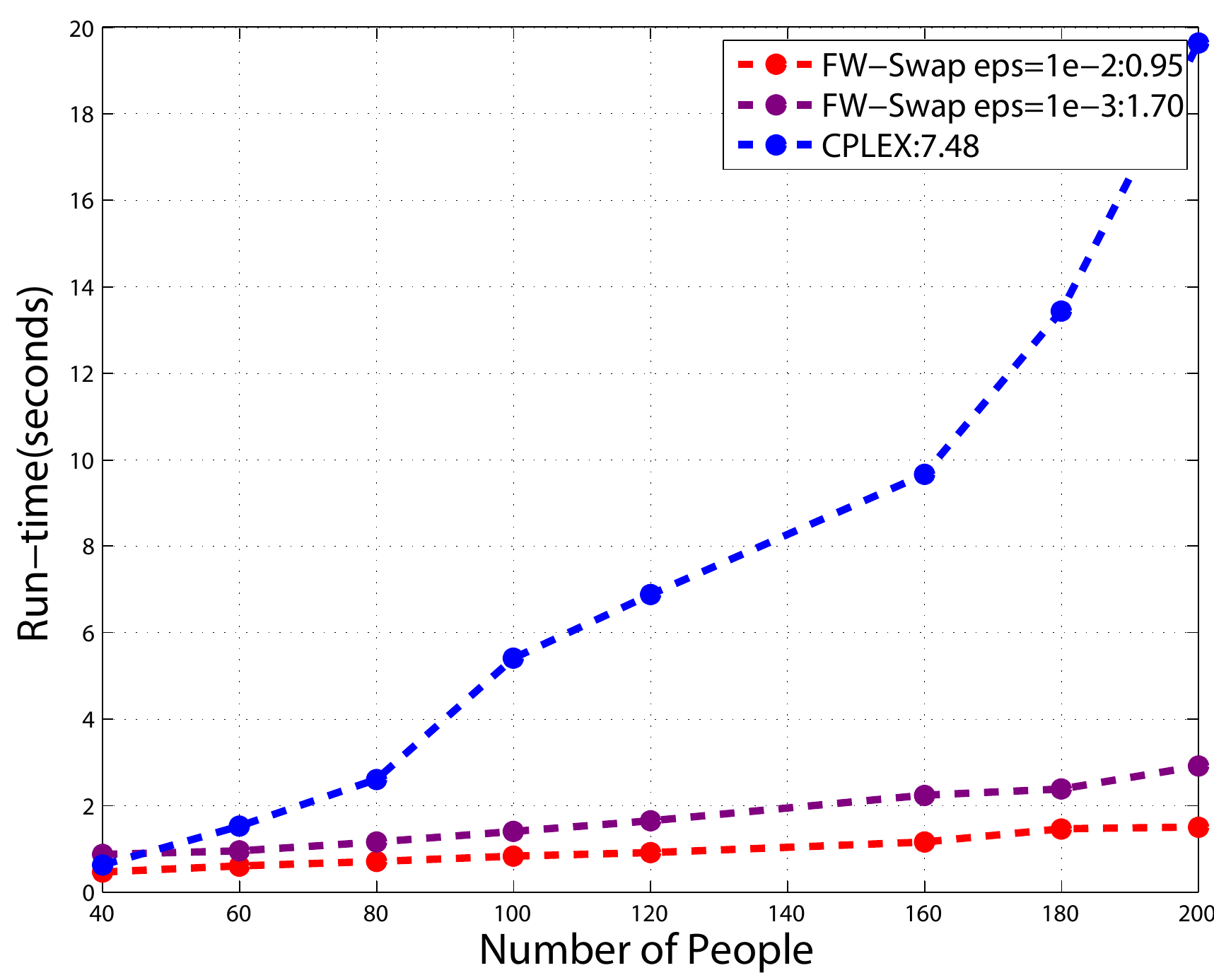}
	\caption{This figure illustrates the run-time comparison of the proposed method for different values of duality gaps $\varepsilon$. We also compare the run-time of FW-Swap with ILOG CPLEX. It is evident that the FW method scales to much larger problem size and requires far less computations.}
	\label{fig:runTimeCplex}
\end{figure}

\begin{figure}[ht!]
	\centering
	\includegraphics[width=0.65\linewidth]{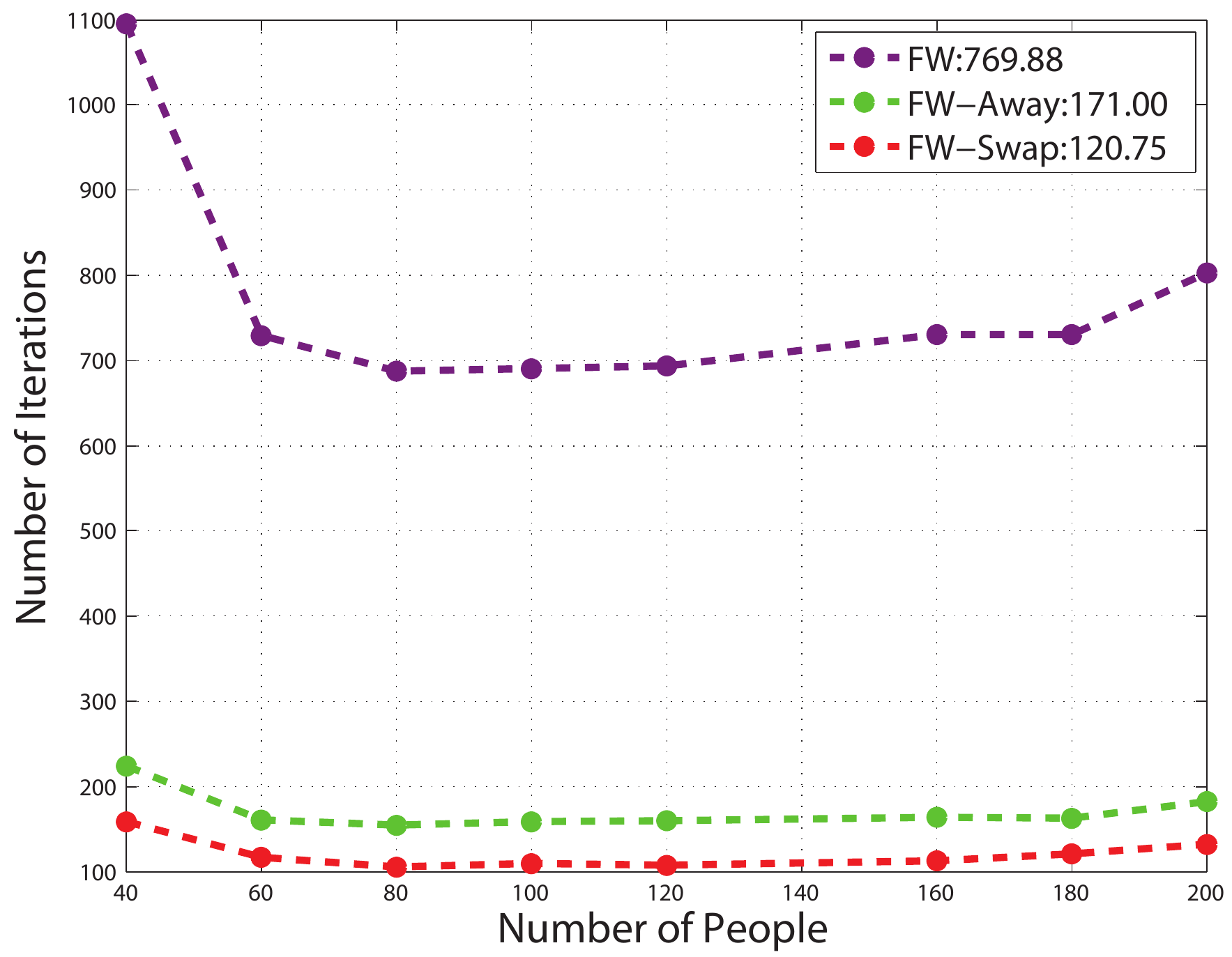}
	\caption{This figures illustrates the number of iterations that it takes for the optimization to converge for original Frank-Wolfe (FW), Frank-Wolfe with Away steps (FW-Away) and Frank-Wolfe with Swap steps (FW-Swap). The value of $\varepsilon$ is set to $0.0001$. }
	\label{fig:iteration-convergence}
\end{figure}

\subsection{Run-time Comparison}
We conduct several experiments to compare run-time of our method with competitive optimization methods. Firstly, we provide a runtime comparison of the FW-Swap in crowd tracking with previous versions of FW, including the original FW \cite{FW} and widely used version of FW with Away steps \cite{JoulinVidLocECCV14}. The results, tested on one of our sequences, are shown in Figure \ref{fig:runTimeFW}. As can be seen the run-time increases as the number of people increases. The duality-gap which determines the stopping criteria and quality of the final solution is set to $ \varepsilon=0.00001$. In practice, we select a higher number $ \varepsilon=0.01$, however the reason we set it to a lower number is that, in this experiment, we are interested in the convergence of these methods.  

In the second experiment, we compare our optimization with publicly available QP solvers. We selected ILOG CPLEC \cite{CPLEX}, which is one of most popular solvers and is used extensively in research community. We compare the run-time of CPLEX with FW-Swap for different duality gap values (in our experiments we set $\varepsilon = 0.01$). The results are shown in Fig. \ref{fig:runTimeCplex}. It is clear that the complexity of CPLEX, even after relaxing the binary constraint, is still very high as the number of targets increases. Finally, we show the number of iterations that our algorithm takes to converge for different setups in Figure \ref{fig:iteration-convergence}. All the experiments are performed on a quad-core 3.0 GHz machine.

\section{Conclusion}
\label{sec:conclusion}
In this paper, we showed that the multi-target tracking in high density crowded scenes can be formulated through Binary Quadratic Programming. Our formulation includes several components that are important in designing a good tracker that works for crowded scenes. Those components include, appearance, motion, neighborhood motion, pairwise spatial relationship and pairwise group information. We show that the proposed formulation can be efficiently solved using Frank-Wolfe optimization with SWAP steps. Additionally, the proposed speed-up technique can reduce the computational complexity, which is necessary when dealing with large number of people. We tested our algorithm on publicly available sequences as well as new sequences and showed state of the art performance. We hope that our paper opens up the room for other researchers to further study this important yet challenging problem.

\bibliographystyle{IEEEtran}
\bibliography{crowdTracker_v2}

\begin{IEEEbiography}[{\includegraphics[width=1in,height=1.25in,clip,keepaspectratio]{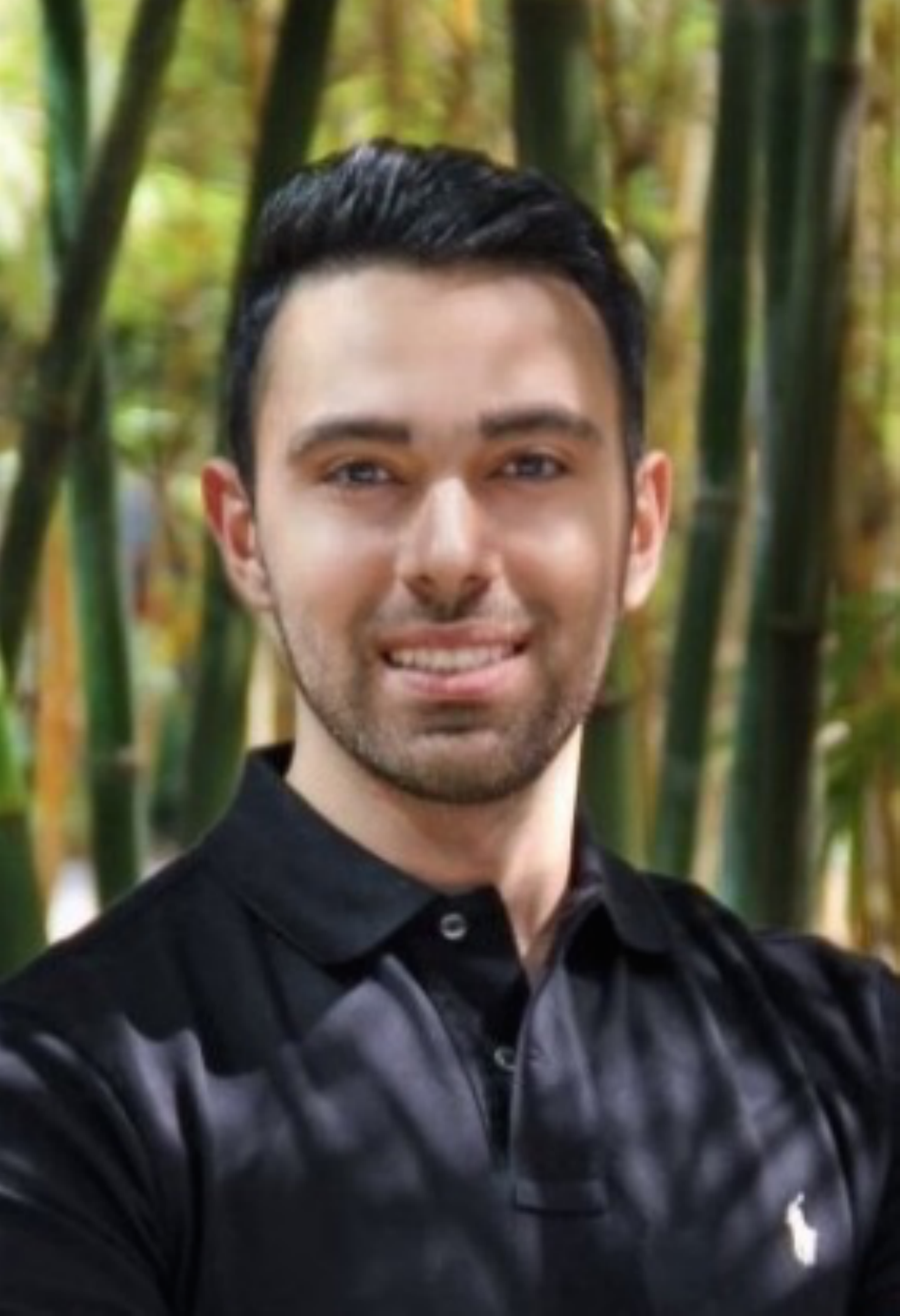}}]{Afshin Dehghan}
received the MS degree in computer science from the University of Central Florida (UCF). He has been with UCF's Center for Research in Computer Vision (CRCV) since 2011, where he is currently working toward the PhD degree in the computer science. He has published several papers in conferences and journals such as CVPR, ECCV, ACM Multimedia, and IEEE Transactions on Pattern Analysis and Machine Intelligence. He was a program committee member of the ACM MM. His research interests include multi-target tracking, complex event recognition, face recognition, mathematical optimization, and graph theory. He received UCF's computer science doctoral student of the year award in 2014-2015.
\end{IEEEbiography}

\begin{IEEEbiography}[{\includegraphics[width=1in,height=1.25in,clip,keepaspectratio]{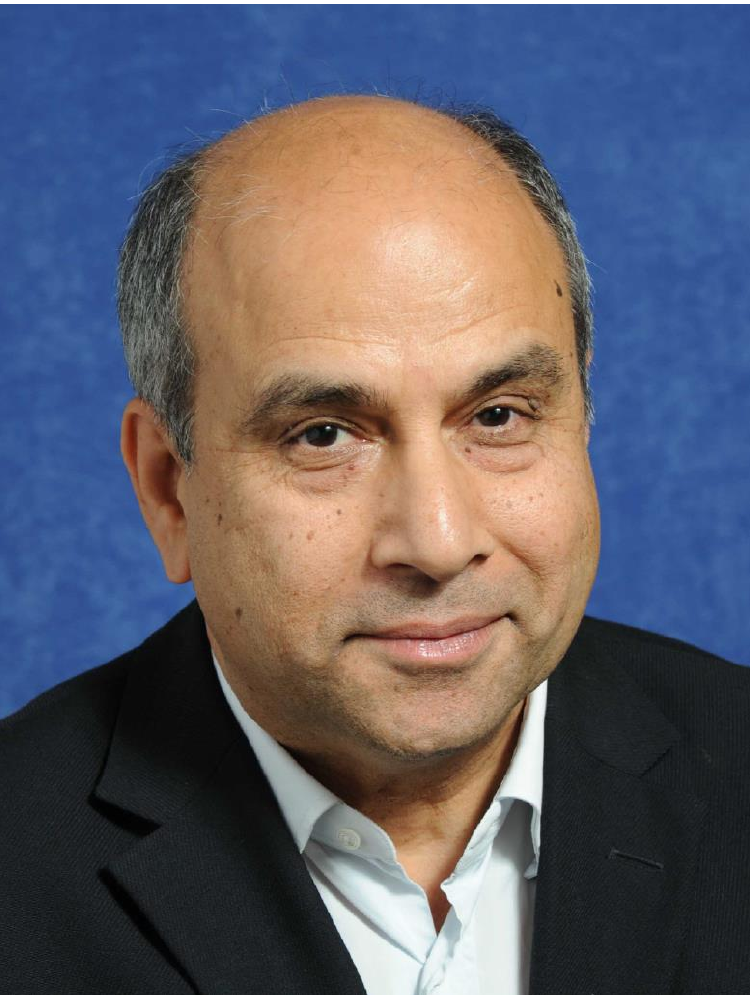}}]{Mubarak Shah}
is the trustee chair professor of computer science and the founding director of the Center for Research in Computer Vision at the University of Central Florida (UCF). He is an editor of an international book series on video computing, editor-in-chief of Machine Vision and Applications Journal, and an associate editor of ACM Computing Surveys Journal. He was the program cochair of the IEEE Conference on Computer Vision and Pattern Recognition (CVPR) in 2008, an associate editor of the IEEE Transactions on
Pattern Analysis and Machine Intelligence, and a guest editor of the special issue of the International Journal of Computer Vision on Video Computing. His research interests include video surveillance, visual tracking, human activity recognition, visual analysis of crowded scenes, video registration, UAV video analysis, and so on. He is an ACM distinguished speaker. He was an IEEE distinguished visitor speaker for 1997-2000 and received the IEEE Outstanding Engineering Educator Award in 1997. In 2006, he was awarded a Pegasus Professor Award, the highest award at UCF. He received the Harris Corporation's Engineering Achievement Award in 1999, TOKTEN Awards from UNDP in 1995, 1997, and 2000, Teaching Incentive Program Award in 1995 and 2003, Research Incentive Award in 2003 and 2009, Millionaires Club Awards in 2005 and 2006, University Distinguished Researcher Award in 2007, Honorable mention for the ICCV 2005 Where Am I? Challenge Problem, and was nominated for the Best Paper Award at the ACM Multimedia Conference in 2005. He is a fellow of the IEEE, AAAS, IAPR, and SPIE.
\end{IEEEbiography}



\end{document}